\documentclass{article}

\usepackage{microtype}
\usepackage{graphicx}
\usepackage{booktabs} % for professional tables

\usepackage{hyperref}

\usepackage{placeins}
\usepackage{url}
\usepackage{booktabs}       % professional-quality tables
\usepackage{amsfonts}       % blackboard math symbols
\usepackage{nicefrac}       % compact symbols for 1/2, etc.
\usepackage{microtype}      % microtypography
\usepackage{graphicx}
\usepackage{siunitx,etoolbox,booktabs}
\usepackage{tabularx}
\usepackage{multirow}
\usepackage{subcaption}
\usepackage{caption}
\usepackage{amsmath}
\usepackage{amsthm}
\usepackage{amssymb}
\usepackage{appendix}
\usepackage{enumitem}
\newlist{steps}{enumerate}{1}
\setlist[steps, 1]{label = Step \arabic*:}
\usepackage[accepted]{icml2021}

\icmltitlerunning{Selfish Sparse RNN Training}

\begin{document}

\twocolumn[
\icmltitle{Selfish Sparse RNN Training}




\begin{icmlauthorlist}
\icmlauthor{Shiwei Liu}{tue}
\icmlauthor{Decebal Constantin Mocanu}{tue,UT}
\icmlauthor{Yulong Pei}{tue}
\icmlauthor{Mykola Pechenizkiy}{tue}
\end{icmlauthorlist}

\icmlaffiliation{tue}{Department of Computer Science, Eindhoven University of Technology, the Netherlands\\}
\icmlaffiliation{UT}{Faculty of Electrical Engineering, Mathematics, and Computer Science at University of Twente, the Netherlands\\}

\icmlcorrespondingauthor{Shiwei Liu}{s.liu3@tue.nl}
% \icmlcorrespondingauthor{Eee Pppp}{ep@eden.co.uk}

\icmlkeywords{Machine Learning, ICML}

\vskip 0.3in
]



\printAffiliationsAndNotice{}  % leave blank if no need to mention equal contribution

\begin{abstract}
Sparse neural networks have been widely applied to reduce the computational demands of training and deploying over-parameterized deep neural networks. For inference acceleration, methods that discover a sparse network from a pre-trained dense network (dense-to-sparse training) work effectively. Recently, dynamic sparse training (DST) has been proposed to train sparse neural networks without pre-training a dense model (sparse-to-sparse training), so that the training process can also be accelerated. However, previous sparse-to-sparse methods mainly focus on Multilayer Perceptron Networks (MLPs) and Convolutional Neural Networks (CNNs), failing to match the performance of dense-to-sparse methods in the Recurrent Neural Networks (RNNs) setting. In this paper, we propose an approach to train intrinsically sparse RNNs with a fixed parameter count in one single run, without compromising performance. During training, we allow RNN layers to have a non-uniform redistribution across cell gates for better regularization. Further, we propose SNT-ASGD, a novel variant of the averaged stochastic gradient optimizer, which significantly improves the performance of all sparse training methods for RNNs. Using these strategies, we achieve state-of-the-art sparse training results, better than the dense-to-sparse methods, with various types of RNNs on Penn TreeBank and Wikitext-2 datasets. Our codes are available at~\url{https://github.com/Shiweiliuiiiiiii/Selfish-RNN}.

%We further investigate a method to measure the topological distance between two sparse neural networks. We find that Selfish-RNN optimizes the same-initialized random topology to completely different optimal topologies at different runs. 
\end{abstract}

\section{Introduction}

Recurrent neural networks (RNNs) \citep{elman1990finding}, with a variant of long short-term memory (LSTM) \citep{hochreiter1997long}, have been highly successful in various fields, including language modeling \citep{mikolov2010recurrent}, machine translation \citep{kalchbrenner2013recurrent}, question answering \citep{hirschman1999deep,wang2017machine}, etc. As a standard task to evaluate models' ability to capture long-range context, language modeling has witnessed great progress in RNNs. \citet{mikolov2010recurrent} demonstrated that RNNs perform much better than backoff models for language modeling. After that, various novel RNN architectures such as Recurrent Highway Networks (RHNs) \citep{zilly2017recurrent}, Pointer Sentinel Mixture Models \citep{merity2017pointer}, Neural Cache Model \citep{grave2016improving}, Mixture of Softmaxes (AWD-LSTM-MoS) \citep{yang2018breaking}, Ordered Neurons LSTM (ON-LSTM) \citep{shen2018ordered}, and effective regularization like Variational Dropout \citep{gal2016theoretically}, Weight Tying \citep{inan2016tying}, DropConnect \citep{merity2018regularizing} have been proposed to improve the performance of RNNs on language modeling.
\begin{figure*}[ht]
\begin{center}
\vspace{-0.1em}
\centerline{\includegraphics[width=16cm, height=4cm]{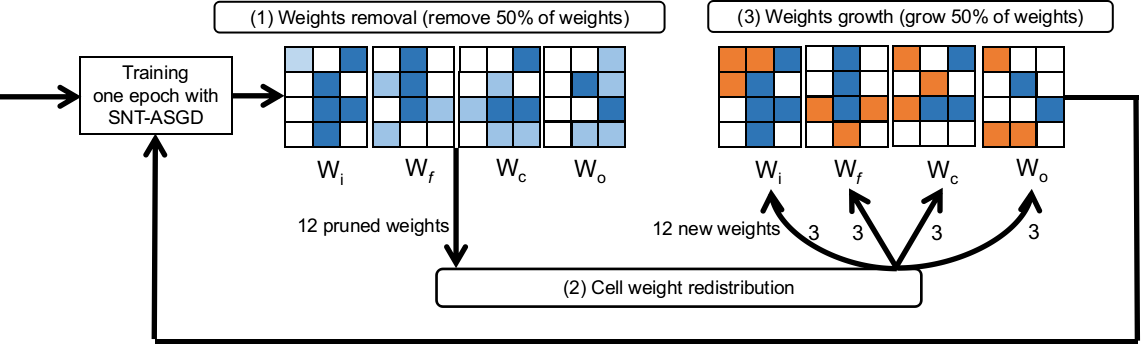}}
\caption{Schematic diagram of the Selfish-RNN. $W_i, W_f, W_c, W_o$ refer to LSTM cell gates. Colored squares and white squares refer to nonzero weights and zero weights, respectively. Light blue squares are weights to be removed and orange squares are weights to be grown.}
\label{Fig:selfishRNN}
\end{center}
\vspace{-0.1em}
\end{figure*}

At the same time, as the performance of deep neural networks (DNNs) improves, the resources required to train and deploy these deep models are becoming prohibitively large. To tackle this problem, various dense-to-sparse methods have been developed, including but not limited to pruning \citep{lecun1990optimal,han2015learning,molchanov2016pruning}, variational dropout \citep{kingma2015variational,molchanov2017variational}, distillation \citep{hinton2015distilling}, $L_1$ regularization \citep{wen2018learning}, and low-rank decomposition \citep{jaderberg2014speeding}. 
Given a pre-trained model, these methods work effectively to accelerate the inference process. 

Recently, some dynamic sparse training (DST) approaches \citep{mocanu2018scalable,mostafa2019parameter,dettmers2019sparse,evci2019rigging} have been proposed to bring efficiency to the training phase as well. However, previous approaches are mainly for CNNs and MLPs. The long-term dependencies and repetitive usage of recurrent cells make RNNs more difficult to be sparsified \citep{kalchbrenner2018efficient,evci2019rigging}. More importantly, the state-of-the-art performance achieved by RNNs on language modeling is mainly associated with the optimizer, averaged stochastic gradient descent (ASGD) \citep{polyak1992acceleration}, which is not compatible with the existing DST approaches. The abovementioned problems heavily limit the performance of the off-the-shelf sparse training methods in the RNN field. For instance, while ``The Rigged Lottery” (RigL) achieves state-of-the-art sparse training results with various CNNs, it fails to match the performance of the iterative pruning method \citep{gale2019state} in the RNN setting \citep{evci2019rigging}.

In this paper, we propose an algorithm to train initially sparse RNNs with a fixed number of parameters throughout training. We abbreviate our sparse RNN training method as Selfish-RNN because our method encourages cell gates to obtain their parameters selfishly. The main contributions of this work are four-fold:
% Furthermore, as the total parameter counts of the sparse RNNs is constrained to a small fraction of dense networks, the corresponding benefits of sparse neural networks, such as computational efficiency and lower memory request, will be obtained in both, training phase and inference phase.  Concretely, we allow dynamical redistribution of cell gates during training. We also propose a new optimizer, sparse non-monotonically triggered averaged stochastic gradient descent (SNT-ASGD), to avoid the sparse structure being damaged by the iterative average operation of the standard NT-ASGD. The ablation results show that this optimizer is the key to reach state-of-the-art RNNs performance.
\begin{itemize}
  \item We propose an algorithm to train sparse RNNs from scratch with a fixed number of parameters. Our method has two novelty components: (1) we propose SNT-ASGD, a sparse variant of the non-monotonically triggered averaged stochastic gradient descent optimizer (NT-ASGD), which improves the performance of all sparse training methods for RNNs; (2) we allow RNN layers to have a non-uniform redistribution across cell gates during training for a better regularization.
  \item We demonstrate state-of-the-art sparse training performance, better than the dense-to-sparse methods, with various RNN models, including stacked LSTMs \citep{zaremba2014recurrent}, RHNs, ON-LSTM on Penn TreeBank (PTB) dataset \citep{marcus1993building} and AWD-LSTM-MoS on WikiText-2 dataset \citep{melis2018on}.
  \item  We present an approach to measure the topological distance between different sparse connectivities from the perspective of graph theory. Recent works \citep{garipov2018loss,draxler2018essentially,fort2019large} on understanding dense loss landscapes find that many independent optima are located in different low-loss tunnels. We complement these works by showing that there exist many low-loss sparse networks which are very different in the topological space.
  \item Our analysis shows two surprising phenomena in the setting of RNNs contrary to CNNs (1) random-based weight growth performs better than gradient-based weight growth, and (2) uniform sparse distribution performs better than \textit{Erd{\H{o}}s-R{\'e}nyi} (ER) sparse distribution. These results highlight the need to choose different sparse training methods for different architectures. 
\end{itemize}

\section{Related Work}
\label{relatedwork}
%\subsection{RNNs Regularization Advances}
%\textcolor{blue}{It is well-known that RNNs, especially the most commonly-used RNN variant, LSTMs, tend to overfit due to the complicated architectures and}, therefore, it is also difficult for normal regularizers like dropout \citep{srivastava2014dropout} to be applied directly to RNNs \citep{zaremba2014recurrent}. Based on insights from variational Bayesian inference, \citet{gal2016theoretically} proposed a variant of LSTM, known as Variational LSTM, which only samples binary dropout masks once for the whole time steps, achieving consistent performance improvement. Weight tying \citep{inan2016tying} significantly compresses the model size by tying the input embedding and output layer together. Rather than operating dropout on the RNN's hidden state vector or the update to the memory state, \citet{merity2017regularizing} proposed DropConnect applying dropout to the recurrent hidden weight tensors to release overfitting on the recurrent connections. Besides, another type of regularization, batch normalization \citep{cooijmans2016recurrent} was also brought to hidden-to-hidden transition to reduce the internal covariate shift between time steps. 

%\subsection{Sparse Training Approaches}
%\label{STT}
\begin{table*}[ht!]
\centering
\vspace{-0.1em}
\caption{Comparison of different sparsity-inducing approaches in RNNs. ER and ERK refer to the \textit{Erd{\H{o}}s-R{\'e}nyi} distribution and the \textit{Erd{\H{o}}s-R{\'e}nyi-Kernel} distribution, respectively. \textit{Backward Sparse} means a clean sparse backward pass and no need to compute or store any information of the non-existing weights. \textit{Sparse Opt} indicates whether a specific optimizer is proposed for sparse RNN training.}
\label{tab:com_different_app}
\begin{tabular}{l|cccccc}
\toprule
Method & Initialization & Removal & Growth & Weight Redistribution &  Backward Sparse & Sparse Opt \\ 
\midrule
Iterative Pruning & dense &  $min(\left| \theta \right|)$ & none & no & no & no\\
ISS & dense & Lasso & none & no & no & no\\
SET & ER &$min(\left| \theta \right|)$ & random & no & yes & no\\
DSR & uniform & $min(\left| \theta \right|)$ & random & across all layers  & yes & no\\
SNFS & uniform & $min(\left| \theta \right|)$ & momentum & across all layers & no & no\\
RigL & ERK &$min(\left| \theta \right|)$ & gradient & no & no & no\\
Selfish-RNN &uniform &$min(\left| \theta \right|)$ & random &  across RNN cell gates & yes & yes\\
\bottomrule
\end{tabular}
\vspace{-0.1em}
\end{table*}
\textbf{Dense-to-Sparse.} There are a large number of works operating on a dense network to yield a sparse network. We divide them into three categories based on the training cost in terms of memory and computation. 
\\(1) \textit{Iterative Pruning and Retraining.} To the best of our knowledge, pruning was first proposed by \citet{janowsky1989pruning} and \citet{mozer1989using} whose goal is to yield a sparse network from a pre-trained network for sparse inference. \citet{han2015learning} brought it back to people's attention based on the idea of iterative pruning and retraining with modern architectures. Gradual Magnitude Pruning (GMP) \citep{zhu2017prune,gale2019state} was further proposed to obtain the target sparse model in one running. Recently, \citet{frankle2018the} proposed the Lottery Ticket Hypothesis showing that the sub-networks (``winning tickets'') obtained via iterative magnitude pruning combined with their ``lucky'' initialization can outperform the dense networks. \citet{zhou2019deconstructing} found that the sign of the initial weights is the key factor that makes the ``winning tickets" work. Our work shows that there exists a much more efficient and robust way to find those ``winning tickets'' without any pre-training steps and any specific initialization. Overall, iterative pruning and retraining requires at least the same training cost as training a dense model, sometimes even more, as a pre-trained dense model is involved. We compare our method with the state-of-the-art pruning method proposed by \citet{zhu2017prune} in Appendix \ref{com_pruning}. With fewer training costs, our method is able to discover sparse networks with lower test perplexity. 
\\(2) \textit{Learning Sparsity During Training.} Some works attempt to learn the sparse networks during training. \citet{louizos2017learning} and \citet{wen2018learning} were examples that gradually enforce the network weights to zero via $L_0$ and $L_1$ regularization, respectively. \citet{dai2018fast} proposed the idea of using singular value decomposition (SVD) to accelerate the training process for LSTMs. Recent works \citep{LIU2020Dynamic,srivastava2015training,xiao2019autoprune,kusupati2020soft} induce sparsity by jointly learning masks and model weights during training. These methods start with a fully dense network, and hence are not memory efficient.
\\(3) \textit{Pruning at Initialization.} Some works aim to find sparse neural networks by pruning once prior to the main training phase based on some salience criteria, including connection sensitivity \citep{lee2018snip,Lee2020A}, synaptic flow \citep{tanaka2020pruning}, gradient signal preservation \citep{Wang2020Picking}, and iterative pruning \citep{de2020progressive}. These techniques can find sparse networks before the standard training, but at least one iteration of dense training is involved to identify these trainable sparse networks. Additionally, pruning at initialization generally cannot match the performance of dynamic sparse training, especially at extreme sparsity levels \citep{Wang2020Picking}.  

\textbf{Sparse-to-Sparse.} Recently, many works have emerged to train intrinsically sparse neural networks from scratch to obtain efficiency both for training and inference.
 \\(1) \textit{Static Sparse Training.} \citet{Mocanu2016xbm} developed intrinsically sparse networks by exploring the scale-free and small-world topological properties in Restricted Boltzmann Machines. Later, some works focus on designing sparse CNNs based on Expander graphs and show comparable performance against the corresponding dense models \citep{prabhu2018deep,kepner2019radix}. \\(2) \textit{Dynamic Sparse Training.}  \citet{mocanu2018scalable} proposed Sparse Evolutionary Training (SET) allowing sparse training MLPs to match the performance of dense MLPs by dynamically changing the sparse connectivity based on a simple remove-and-regrow strategy. Shortly after, DeepR was introduced by \citet{bellec2018deep} to train sparse networks by sampling the sparse connectivity from the posterior distribution. \citet{mostafa2019parameter} introduced Dynamic Sparse Reparameterization (DSR) to train sparse neural networks while dynamically adjusting the sparse distribution during training. Sparse Networks from Scratch (SNFS) \citep{dettmers2019sparse} improved the sparse training performance by introducing the idea of growing free weights according to their momentum. While effective, it requires extra computation and memory to update the dense momentum tensor for every iteration. RigL \citep{evci2019rigging} went one step further by activating new weights with the highest magnitude gradient. It amortizes the significant amount of overhead by updating the sparse connectivity infrequently. In addition, some recent works \citep{jayakumar2020top,raihan2020sparse} attempt to explore more sparse space during training to improve the sparse training performance. Due to the inherent limitations of deep learning software and hardware libraries, all of those works simulate sparsity using a binary mask over weights. More recently, \citet{liu2020sparse} proved the practical values of DST by developing for the first time an independent software framework to train truly sparse MLPs with over one million neurons on a typical laptop. However, all these works mainly focus on CNNs and MLPs, and they are not designed to match state-of-the-art performance for RNNs.

\begin{figure}[ht!]
\vspace{-0.1em}
\begin{center}
\includegraphics[width=7cm, height=5.1cm]{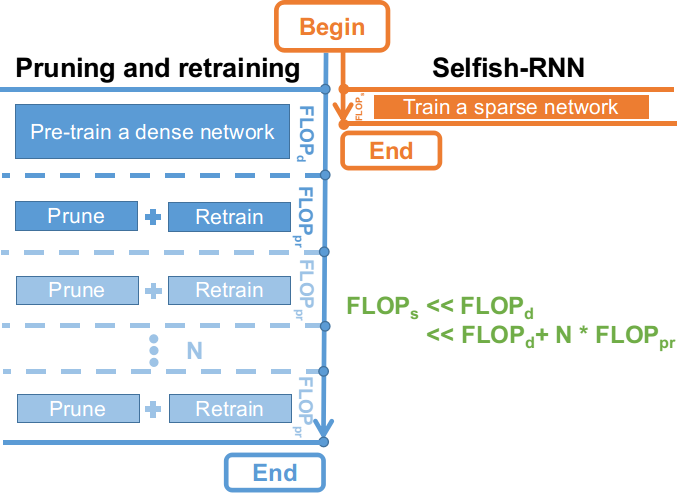}
\caption{ A high-level overview of the difference between Selfish-RNN and iterative pruning and re-training techniques. Blocks with light blue color represent optional pruning and retraining steps chosen depending on specific approaches. }
\label{Fig:Overview}
\end{center}
\vspace{-0.1em}
\end{figure}

We summarize the properties of all approaches compared in this paper in Table \ref{tab:com_different_app}. Additionally, we provide a high-level overview of the difference between Selfish-RNN and iterative pruning and re-training in Figure \ref{Fig:Overview}. FLOPs required by Selfish-RNN is much smaller than iterative pruning and re-training, as it starts with a sparse network and without any retraining phases. See Appendix \ref{diff_set_RNN_pruning} for more differences. 
 
\section{Sparse RNN Training}

Our sparse RNN training method is illustrated in Figure \ref{Fig:selfishRNN} with LSTM as a specific case. Note that our method can be easily applied to any other RNN variants. The only difference is the number of cell gates. Before training, we randomly initialize each layer at the same sparsity (the fraction of zero-valued weights), so that the training costs are proportional to the dense model at the beginning. To explore more sparse structures, while to maintain a fixed sparsity level, we need to optimize the sparse connectivity together with the corresponding weights (a combinatorial optimization problem). We use dynamic sparse connectivity and SNT-ASGD together to handle this combinatorial optimization problem. The pseudocode of the full training procedure of our algorithm is shown in Algorithm \ref{alg:Selfish-RNN}. 
\subsection{Dynamic Sparse Connectivity}
We consider uniform sparse initialization, magnitude weight removal, random weight growth, and cell gate redistribution together as main components of our dynamic sparse training method, which can ensure a fixed number of parameters and a clean sparse backward pass, as discussed next.
\begin{algorithm}[tb]
   \caption{Selfish-RNN}
   \label{alg:Selfish-RNN}
\begin{algorithmic}
   \STATE {\bfseries Input:} Model weight $\theta$, number of layer $L$, sparsity $S$, pruning rate $p$, training epoch $n$
   \FOR{$i=1$ \textbf{to} $L$}
   \STATE Initialize the network with Eq. (\ref{Eq:initialization})
   \ENDFOR
   \FOR{$epoch=1$ \textbf{to} $n$}
    \STATE Training: $\theta_s\leftarrow$ SNT-ASGD$(\theta_s)$ 
        \FOR{$i=1$ \textbf{to} $L$}
            \IF{RNN layer}
            \STATE Cell Gate redistribution with Eq. (\ref{Eq:rnn_pruning}) \;
            \ELSE
            \STATE Weight removal with Eq. (\ref{Eq:pruning}) 
            \STATE Weight growth with Eq. (\ref{Eq:regrow}) \; 
            \ENDIF
        \ENDFOR
        \STATE $p\leftarrow$DecayRemovingRate$(p)$ \;
      \ENDFOR
\end{algorithmic}
\end{algorithm}
\\\textbf{Notation.}
Given a dataset of $N$ samples $\mathbf{D} = \{(x_i, y_i) \}_{i=1}^N$ and a network $f(x; \theta)$ parameterized by $\theta$. We train the network to minimize the loss function $\sum_{i=1}^{N} L(f(x_i;\theta),y_i)$. The basic mechanism of sparse training is to train with a fraction of parameters $\theta_s$, while preserving the performance as much as possible. Hence, a sparse neural network can be denoted as $f_s(x; \theta_s)$ with a sparsity level $S = 1 - \frac{\|\theta_s\|_0}{\|\theta\|_0}$, where $\|\cdot\|_0$ is the $\ell_0$-norm. 
\\\textbf{Uniform Sparse Initialization.}
First, the network is randomly initialized with a uniform sparse distribution in which the sparsity level of each layer is the same $S$. More precisely, the network is initialized by:
\begin{equation}
 \theta_s = \theta \odot M
 \label{Eq:initialization}
\end{equation}
where $\theta$ is a dense weight tensor initialized in a standard way; $M$ is a binary tensor, in which nonzero elements are sampled uniformly based on the sparsity $S$; $\odot$ refers to the Hadamard product. 
\\\textbf{Magnitude Weight Removal.}
For non-RNN layers, we use magnitude weight removal followed by random weight growth to update the sparse connectivity. We remove a fraction $p$ of weights with the smallest magnitude after each training epoch. This step is performed by changing the binary tensor $M$, as follows:
\begin{equation}
 M = M - P
 \label{Eq:pruning}
\end{equation}
where $P$ is a binary tensor with the same shape as $M$, in which the nonzero elements have the same indices with the top-$p$ smallest-magnitude nonzero weights in $\theta_s$, with $||P||_0=p||M||_0$. 
\\\textbf{Random Weight Growth.}
To keep a fixed parameter count, we randomly grow the same number of weights immediately after weight removal, by:
\begin{equation}
 M = M + R
 \label{Eq:regrow}
\end{equation}
where $R$ is a binary tensor where the nonzero elements are randomly located at the position of zero elements of $M$. We choose random growth to get rid of using any information of the non-existing weights, so that both feedforward and backpropagation are completely sparse. It is more desirable to have such pure sparse structures as it enables the possibility of conceiving in the future specialized hardware accelerators for sparse neural networks. Besides, our analysis of growth methods in Section \ref{ana_growth} shows that random growth can explore more sparse structural degrees of freedom than gradient growth, which might be crucial to sparse training.
\\\textbf{Cell Gate Redistribution.}
Our dynamic sparse connectivity differs from previous methods mainly in cell gate redistribution. For non-RNN layers, we use magnitude weight removal followed by random weight growth to update the sparse connectivity. For RNN layers, we use cell gate redistribution to update their sparse connectivities. The naive approach is to sparsify all cell gates independently at the same sparsity, as used in \citet{liu2019intrinsically} which is a straightforward SET extension to RNNs. Essentially, it is more desirable to redistribute new weights to cell gates dependently, as all gates collaborate together to regulate information. Intuitively, we redistribute new weights in a way that cell gates containing more large-magnitude weights should have more weights. Large-magnitude weights indicate that their loss gradients are large and few oscillations occur. Therefore, gates with more large-magnitude weights should be reallocated with more parameters to accelerate training. Concretely, for each RNN layer $l$, we remove weights dependently given by an ascending sort:
\begin{equation}
Sort_p(|\theta_1^l|, |\theta_2^l|,.., |\theta_t^l|)
\label{Eq:rnn_pruning}
\end{equation}
where $\{\theta_1^l,\theta_2^l,...,\theta_t^l\}$ are all gate tensors within RNN cell, and $Sort_p$ returns $p$ indices of the smallest-magnitude weights. After weight removal, new weights are uniformly grown to each gate, so that gates with more large-magnitude weights will gradually obtain more weights.  We further demonstrate the final sparsity breakdown of cell gates learned by our method in Appendix \ref{breakdown} and observe that the forget gates are consistently sparser than other gates for all models. 

% Note that redistributing parameters across cell gates does not change the FLOP counting, as the sparsity of each layer is not changed. In contrast, the across-layer weight redistribution used by DSR and SNFS affects the sparsity level of each layer. As a result, it will change the number of floating-point operations (FLOPs). 

Similar with SNFS, We also decay the pruning rate $p$ to zero with a cosine annealing. We further use Eq. (\ref{Eq:initialization}) to enforce the sparse structure before the forward pass and after the backward pass, so that zero weights will not contribute to the loss. And all newly activated weights are initialized to zero.

\begin{figure}[ht]
\vspace{-0.1cm}
\begin{center}
\centerline{\includegraphics[width=7cm, height=4.5cm]{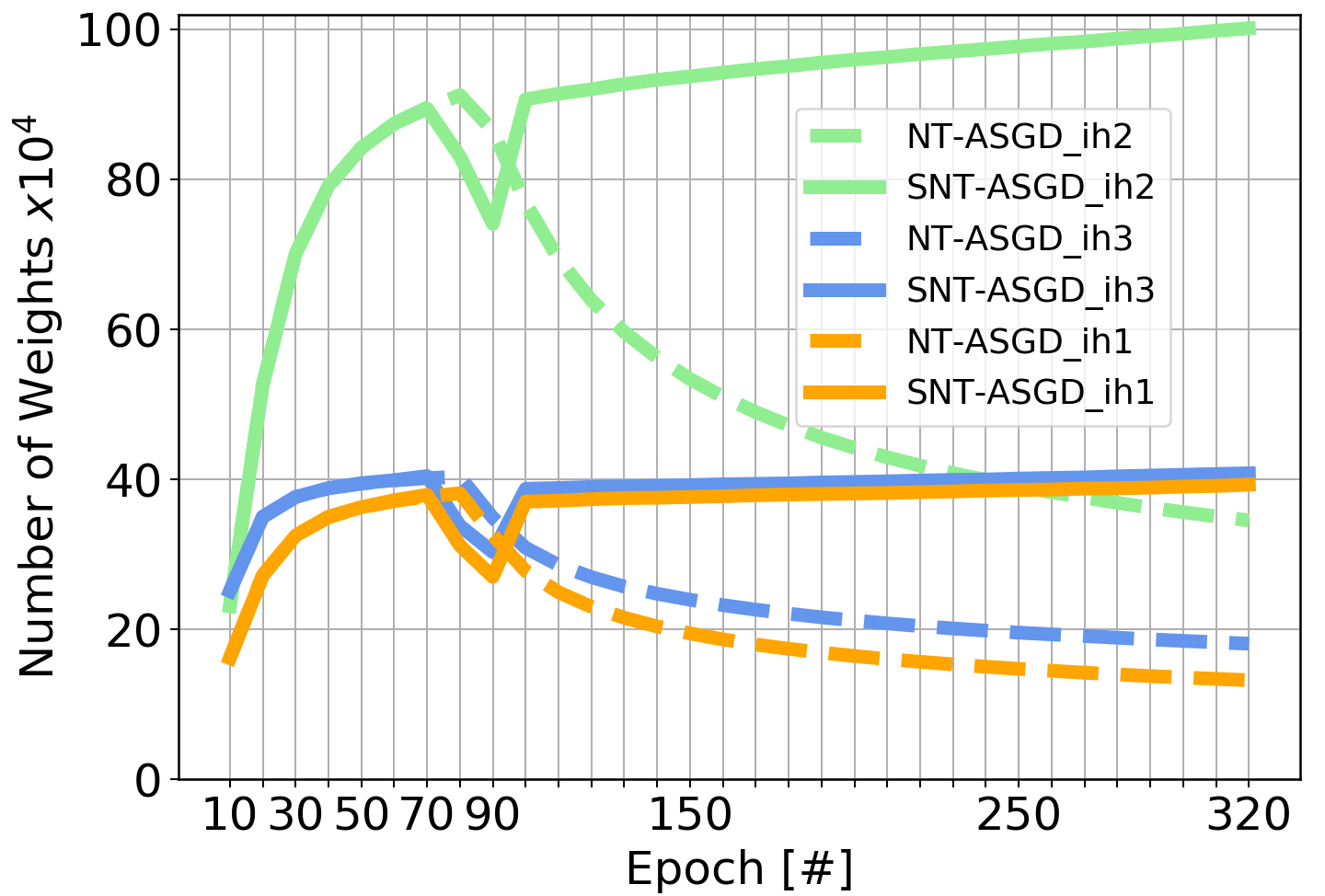}}
\caption{The number of weights whose magnitude is larger than 0.1 during training for ON-LSTM. The solid lines represent SNT-ASGD and  dashed lines represent standard NT-ASGD. \textit{ih1}, \textit{ih2}, and \textit{ih3} refer to input weights in the first, second and third LSTM layer.}
\label{Fig:NumberOfWeights}
\end{center}
\vskip -0.2in
\end{figure}

% \end{minipage}
\subsection{Sparse Non-monotonically Triggered ASGD}

Non-monotonically Triggered ASGD (NT-ASGD) has shown its ability to achieve remarkable performance with various RNNs \citep{merity2018regularizing,yang2018breaking,shen2018ordered}. However, it becomes less appealing for sparse RNNs training. Unlike dense networks in which every parameter in the model is updated at each iteration, the non-active weights remain zero for sparse training. Once these non-active weights are activated, the original averaging operation of standard NT-ASGD will immediately bring them close to zero. Thereby, after the averaging operation is triggered, the number of valid weights will decrease sharply. To alleviate this problem, we describe SNT-ASGD as following:
% \vspace{-0.3cm}
\begin{equation}
\label{Eq:SparseASGD}
\tilde{w_i}=\left\{
\begin{array}{lcl}
0  &   & {if \; m_i=0 , \forall i}, \\
\frac{\sum_{t=T_i}^{K} w_{i, t}}{(K-T_i+1)} & & {if \; m_i=1, \forall i}.
\end{array}
\right.
% \vspace{-0.3cm}
\end{equation}
where $\tilde{w_i}$ is the value returned by SNT-ASGD for weight $w_i$; $w_{i, t}$ represents the actual value of weight $w_i$ at the $t^{th}$ iteration; $m_i = 1$ if the weight $w_i$ exists and $m_i = 0$ if not; $T_i$ is the iteration in which the weight $w_i$ grows most recently; and $K$ is the total number of iterations.  We demonstrate the effectiveness of SNT-ASGD in Figure \ref{Fig:NumberOfWeights}. At the beginning when trained with SGD, the number of weights with high magnitude increases fast. However, the trend of NT-ASGD starts to descend significantly once the averaging is triggered at epoch 80, whereas the trend of SNT-ASGD continues to rise after a small drop caused by the averaging operation. 
\begin{table*}[t]
\vspace{-0.2cm}
\caption{Single model perplexity on validation and test sets for the Penn Treebank language modeling task with stacked LSTMs and RHNs. ‘*’ indicates the reported results from the original papers: ``Dense'' is obtained from \citet{zaremba2014recurrent} and ISS is obtained from \citet{wen2018learning}. ``Static-ER'' and ``Static-Uniform'' are the static sparse network trained from scratch with ER distribution and uniform distribution, respectively. ``Small Dense'' refers to the small dense network with the same number of parameters as Selfish-RNN.}
\label{tab:stackedlstm}
\vskip 0.15in
\begin{center}
\begin{small}
\begin{sc}
\begin{tabular}{l|ccc||ccc}
\toprule
& \multicolumn{3}{c||}{Stacked LSTMs} &  \multicolumn{3}{c}{RHNs}\\
\midrule
Models & \#Parameters & Validation & Test & \#Parameters & Validation & Test \\
\midrule
Dense* & 66.0M & 82.57 & 78.57 & 23.5M  &67.90 & 65.40\\
% & (3.1e16) & (7.2e10) & & &  (6.5e16) & (3.3e10) & &\\
Dense (NT-ASGD) & 66.0M  & 74.51 & 72.40  & 23.5M &  63.44 & 61.84\\
\midrule
\midrule
& \multicolumn{3}{c||}{Sparsity=0.67} &  \multicolumn{3}{c}{Sparsity=0.53}\\
\midrule
Small Dense (NT-ASGD)  & 21.8M & 88.67 & 86.33  & 11.1M & 70.10 & 68.40  \\
Static-ER (SNT-ASGD)&21.8M &  81.02 & 79.30  & 11.1M & 75.74 & 73.21 \\
Static-Uniform (SNT-ASGD)& 21.8M & 80.37 & 78.61 &  11.1M &74.11 & 71.83 \\
\midrule
ISS* & 21.8M &  82.59 & 78.65 & 11.1M &  68.10 & 65.40 \\
DSR (Adam) & 21.8M &  89.95 & 88.16& 11.1M &  65.38 & 63.19 \\
GMP (Adam) & 21.8M &  89.47 & 87.97 & 11.1M& 63.21 & 61.55 \\
SNFS (Adam)  & 21.8M&  88.31 & 86.28 & 11.1M &  74.02 & 70.99 \\
RigL (Adam) & 21.8M & 88.39 & 85.61  & 11.1M&  67.43 & 64.41 \\
SET (Adam) & 21.8M &  87.30 & 85.49& 11.1M &  63.66 & 61.08 \\
Selfish-RNN (Adam) & 21.8M&  85.70 & 82.85 & 11.1M& 63.28 & 60.75\\
RigL (SNT-ASGD) & 21.8M & 78.31 & 75.90 & 11.1M &  64.82 & 62.47 \\
GMP (SNT-ASGD) & 21.8M &  76.78 & 74.84 & 11.1M&  65.63 & 63.96 \\
Selfish-RNN (SNT-ASGD) & 21.8M &   \textbf{73.76} &  \textbf{71.65} & 11.1M & \textbf{62.10} & \textbf{60.35}  \\
\midrule
\midrule
& \multicolumn{3}{c||}{Sparsity=0.62} &  \multicolumn{3}{c}{Sparsity=0.68}\\
\midrule
ISS*  & 25.2M & 80.24 & 76.03  & 7.6M &  70.30 & 67.70  \\
% GMP  & 0.66x & 0.38x & 89.28 & 87.79  & 0.51x & 0.32x & 63.22 &  61.39 \\
RigL (SNT-ASGD) & 25.2M  & 77.16 & 74.76  & 7.6M &  69.32 & 66.64 \\
GMP (SNT-ASGD) & 25.2M & 74.86 & 73.03 & 7.6M &  66.61 & 64.98 \\
Selfish-RNN (SNT-ASGD) & 25.2M & \textbf{73.50} & \textbf{71.42} & 7.6M  &  \textbf{66.35} & \textbf{64.03}  \\
% DSR (SNT-ASGD)  & 25.2M & 0.380x & 0.376x & 72.98 & 71.36 \\
\bottomrule
\end{tabular}
\end{sc}
\end{small}
\end{center}
\vskip -0.1in
\end{table*}

Besides, the constant learning rate of SNT-ASGD helps to prevent the negative effect of learning rate decay on dynamic sparse training. Since most dynamic sparse training methods initialize the newly activated weights as zero, the magnitude of new weights can barely catch up with the unpruned weights once the learning rate is decayed to small values. To better understand how proposed components, cell gate redistribution and SNT-ASGD, improve the sparse RNN training performance, we further conduct an ablation study in Appendix \ref{app:ablation}. It is clear to see that both of them lead to significant performance improvement.
\vspace{-0.3cm}
\section{Experimental Results}
We evaluate Selfish-RNN with various models including stacked LSTMs, RHNs, ON-LSTM on the Penn TreeBank dataset and AWD-LSTM-MoS on the WikiText-2 dataset. The performance of Selfish-RNN is compared with 6 state-of-the-art sparse inducing techniques, including 2  dense-to-sparse methods, ISS and GMP; 4 sparse-to-sparse methods, SET, DSR, SNFS, and RigL. Intrinsic Sparse Structures (ISS) \citep{wen2018learning} is a method that uses Lasso regularization to explore sparsity inside RNNs. GMP is the state-of-the-art unstructured pruning method in DNNs. For fair comparison, we use the exact same hyperparameters and regularization introduced in ON-LSTM \citep{shen2018ordered} and AWD-LSTM-MoS \citep{yang2018breaking}. We then extend these similar settings to stacked LSTMs and RHNs. We choose Adam \cite{kingma2014adam} and SNT-ASGD as the optimizers of different DST methods. Due to the space limitation, we put the results of sparse ON-LSTM on PTB and sparse AWD-LSTM-MoS on Wikitext-2 in Appendix \ref{expre_onlstm} and Appendix \ref{expre_mos}, respectively.  See Appendix \ref{app:Hyperparameters} for hyperparameters.
\subsection{Stacked LSTMs}

As introduced by \citet{zaremba2014recurrent}, stacked LSTMs (large) is a two-layer LSTM model with 1500 hidden units for each LSTM layer. We choose the same sparsity as ISS, 67\% and 62\%.
Results are shown in the left side of Table \ref{tab:stackedlstm} (see Appendix \ref{app:training_flops} for the version with estimated FLOPs). We provide a new dense baseline trained with the standard NT-ASGD, achieving 6 lower test perplexity than the widely-used baseline.  When optimized by Adam, while Selfish-RNN achieves the lowest perplexity, all sparse training techniques fail to match the performance of ISS and dense models. On the other hand, training sparse RNNs with SNT-ASGD substantially improves the performance of all sparse methods, and Selfish-RNN achieves the best one, even better than the new dense baseline. Note that even starting from a sparse network (sparse-to-sparse), our method can discover a better solution than the state-of-the-art dense-to-sparse method, GMP. We also test whether a small dense network and a static sparse network with the same number of parameters as Selfish-RNN can match the performance of Selfish-RNN. We train a dense stacked LSTMs with 700 hidden units, named as ``Small Dense". In line with the previous studies \citep{mocanu2018scalable,mostafa2019parameter,evci2019rigging}, both static sparse networks and small dense networks fall short of Selfish-RNN. Moreover, training a static sparse network from scratch with uniform distribution performs better than the one with \textit{ER} distribution.

\begin{table}[ht]
\vspace{-0.2cm}
\caption{Effect of different optimizers on different DST methods. We choose stacked LSTMs on PTB dataset at a sparsity level of 67\%.}
\label{tab:SNT_STM}
\vspace{-0.3cm}
\begin{center}
\begin{small}
\begin{sc}
\begin{tabular}{lccc}
\toprule
Models & \#Parameters  & Validation & Test \\
\midrule
Dense & 66.0M & 82.57 & 78.57 \\
% Dense (NT-ASGD) & 66.0M & 1x & 1x & 74.51 & 72.40 \\
\midrule
\midrule
& \multicolumn{3}{c}{Adam}\\
\midrule
DSR  & 21.8M & 89.95 & 88.16  \\
SNFS  & 21.8M  & 88.31 & 86.28 \\
RigL   & 21.8M &  88.39 & 85.61 \\
SET  & 21.8M & 87.30 & 85.49 \\
Selfish-RNN & 21.8M  & \textbf{85.70}  &  \textbf{82.85} \\
\midrule
\midrule
& \multicolumn{3}{c}{SGD with Momentum}\\
\midrule
SNFS   & 21.8M & 90.09 & 87.98 \\
SET   & 21.8M & 85.73 & 82.52 \\
RigL  & 21.8M  & 84.78 & 80.81 \\
DSR   & 21.8M& 82.89 & 80.09  \\
Selfish-RNN & 21.8M  &  \textbf{82.48} & \textbf{79.69}  \\
\midrule
\midrule
& \multicolumn{3}{c}{SNT-ASGD}\\
\midrule
SNFS  & 21.8M & 82.11 &  79.50 \\
RigL  & 21.8M & 78.31 & 75.90 \\
% SET  (SNT-ASGD) (MP) & 21.8M & 0.330x & 0.328x & 74.37 & 72.89  \\
SET  & 21.8M &  76.78 & 74.84 \\
Selfish-RNN  & 21.8M &  73.76  & 71.65 \\
DSR  & 21.8M & \textbf{72.30} & \textbf{70.76} \\
\bottomrule
\end{tabular}
\end{sc}
\end{small}
\end{center}
\vspace{-1cm}
\end{table}

To understand better the effect of different optimizers on different DST methods, we report the performance of all DST methods trained with Adam, momentum SGD, and SNT-ASGD. For SNFS (SNT-ASGD), we replace momentum of weights with their gradients, as SNT-ASGD does not involve any momentum terms. We use the same hyperparameters for all DST methods. The results are shown in Table \ref{tab:SNT_STM}. It is clear that SNT-ASGD brings significant perplexity improvements to all sparse training techniques. This further stands as empirical evidence that SNT-ASGD is crucial to improve the sparse training performance in the RNN setting. Moreover, compared with other DST methods, Selfish-RNN is quite robust to the choice of optimizers likely due to its simple scheme to update the sparse connectivity. Advanced strategies such as across-layer weight redistribution used in DSR and SNFS, gradient-based weight growth used in RigL and SNFS heavily depend on optimizers. They might work decently for some optimization methods but may not for others.

\subsection{Recurrent Highway Networks}
Recurrent Highway Networks \citep{zilly2017recurrent} is a variant of RNNs allowing RNNs to explore deeper architectures inside the recurrent transition. The results are shown in the right side of Table \ref{tab:stackedlstm}. Again, Selfish-RNN achieves the lowest perplexity with both Adam and SNT-ASGD, better than the dense-to-sparse methods (ISS and GMP). 
Surprisingly, random-based growth methods (SET, DSR, and Selfish-RNN) consistently have the lower perplexity than the gradient-based growth methods (RigL and SNFS). We further analyze the effect of different weight growth methods on DST in Section \ref{ana_growth}.

\section{Analyzing the Performance of Selfish-RNN}

\begin{figure*}[ht!]
\vspace{-0.1cm}
\begin{subfigure}[b]{0.45\textwidth}
\hspace{0.8cm}
\includegraphics[width=8.0cm, height=5cm]{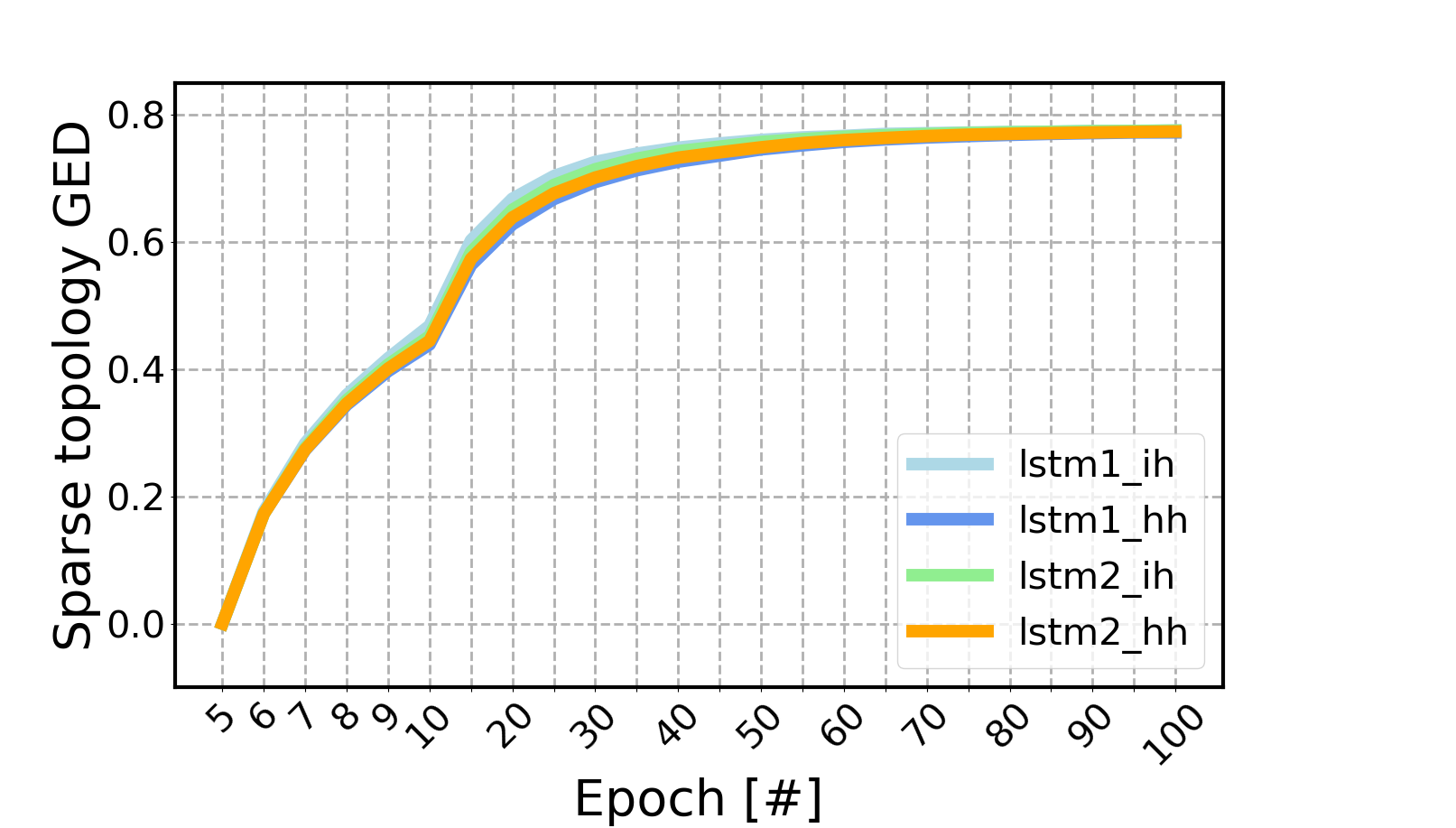}
\end{subfigure}
\hspace{0.7cm}
\begin{subfigure}[b]{0.45\textwidth}
\centering
\includegraphics[width=8.0cm, height=5cm]{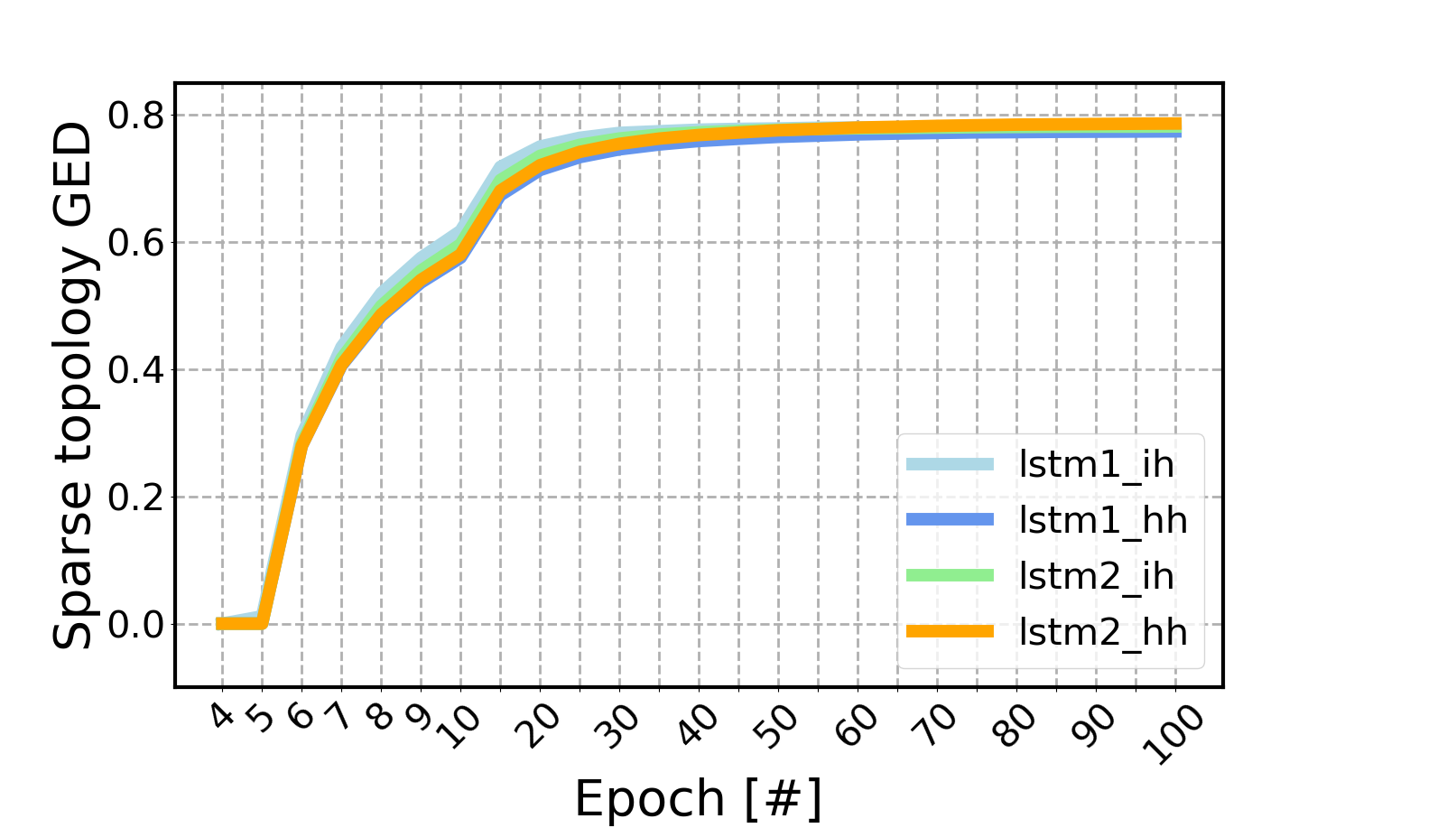}
%   \caption{pruning rate sensitivity}label{fig:pruning_rate}
\end{subfigure}
\caption{(\textbf{left}) One random-initialized sparse stacked LSTMs trained with Selfish-RNN end up with a very different sparse connectivity topology. (\textbf{right}) Two same-initialized sparse stacked LSTMs trained with different random seeds end up with very different sparse connectivity topologies. $ih$ is the input weight tensor comprising four cell gates and $hh$ is the hidden state weight tensor comprising four cell gates.}
\label{fig:GED}
\end{figure*}

\begin{figure*}[h]
\vspace{-0.1cm}
\begin{subfigure}[b]{0.45\textwidth}
\hspace{0.8cm}
  \includegraphics[width=8cm, height=5cm]{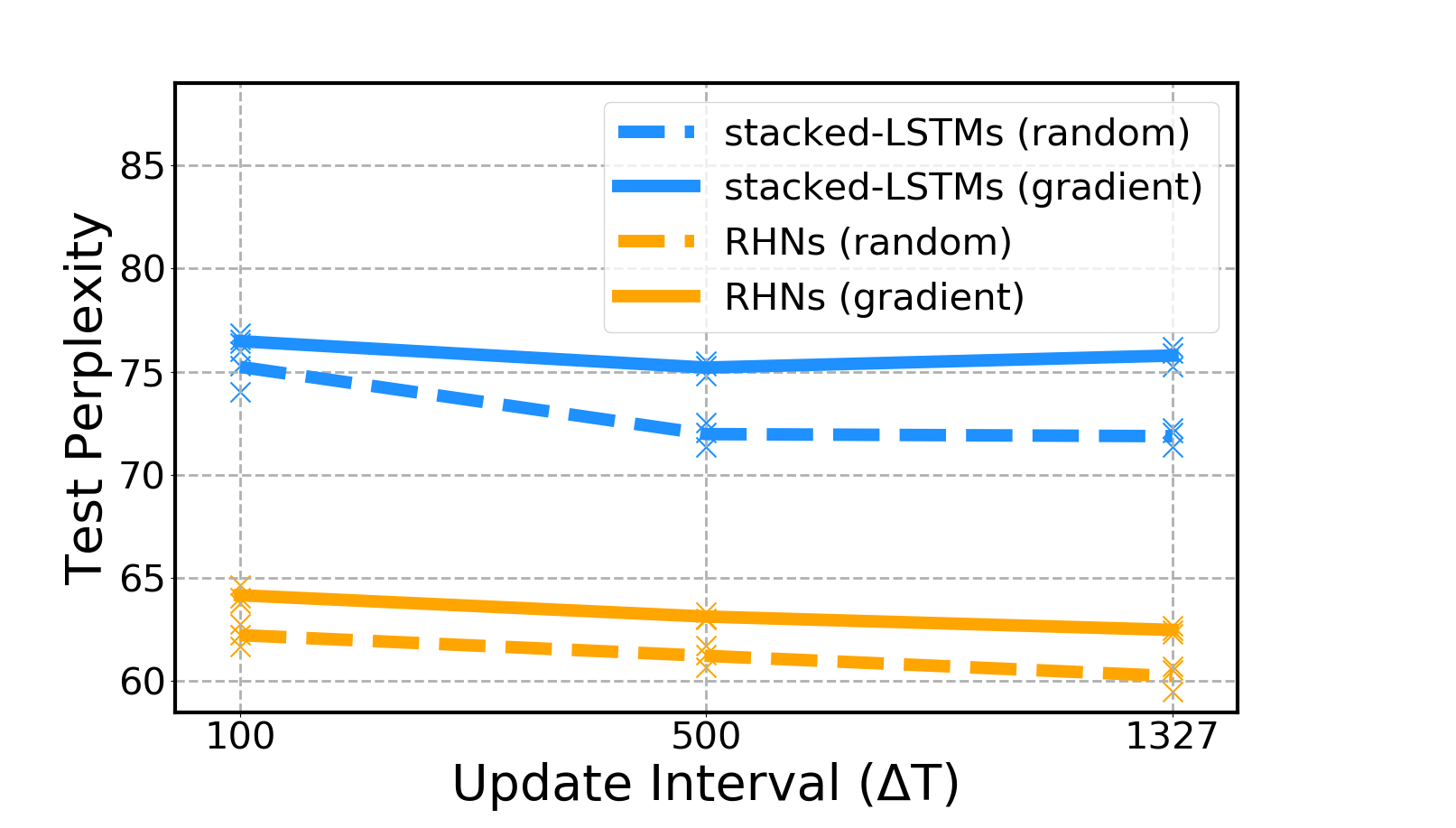}
%   \caption{Sparsity sensitivity}\label{fig:sparsity}
\end{subfigure}
\hspace{0.7cm}
\begin{subfigure}[b]{0.45\textwidth}
  \includegraphics[width=8cm, height=5cm]{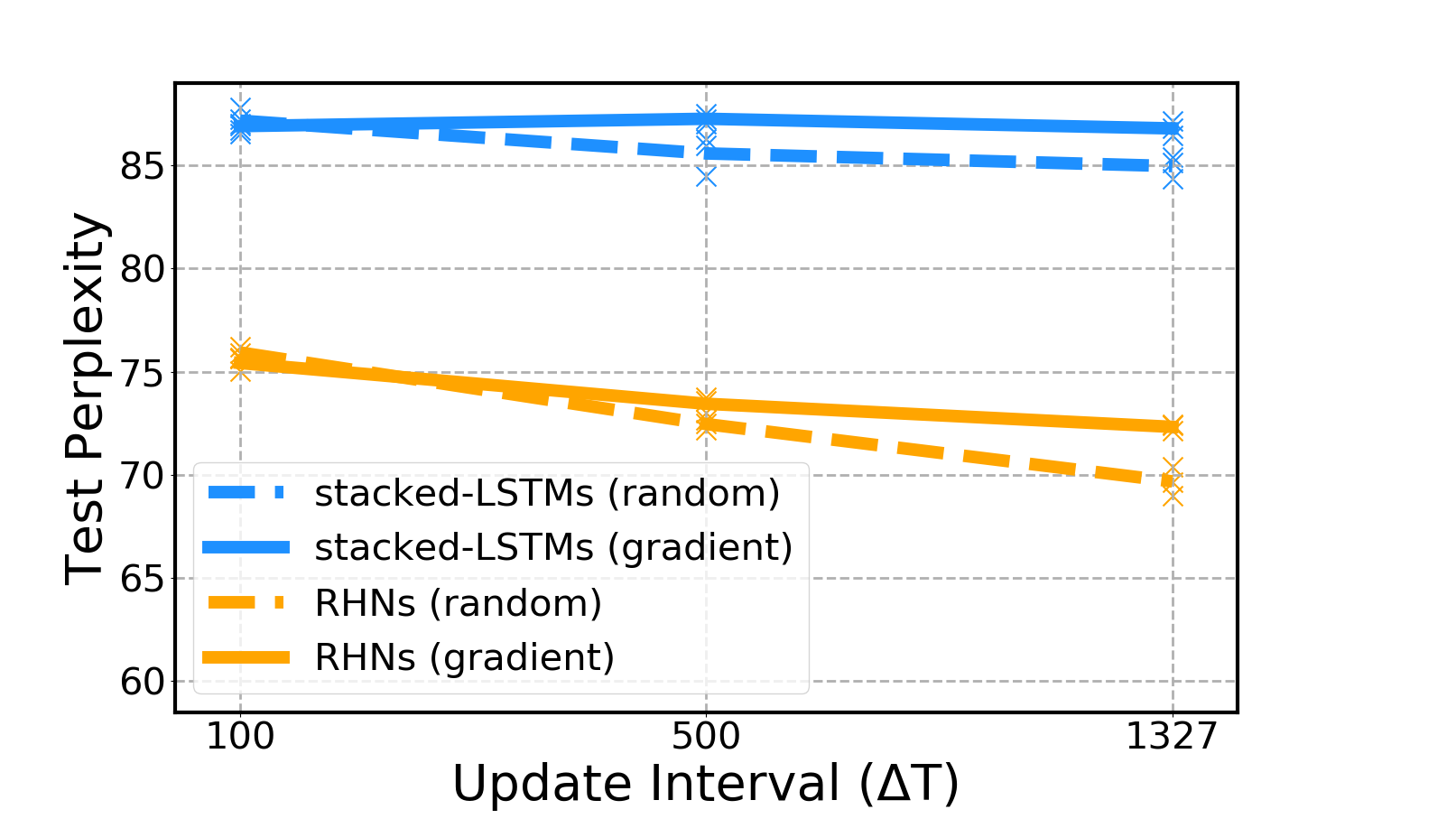}
%   \caption{pruning rate sensitivity}\label{fig:pruning_rate}
\end{subfigure}\hfill
% \vspace{-0.2cm}
\caption{Comparison between random-based growth and gradient-based growth. (\textbf{left}) Models trained with SNT-ASGD. (\textbf{right}) Models trained with momentum SGD. }
\label{fig:com_gra_ran}
\end{figure*}

\subsection{Analysis of Topological Distance between Sparse Connectivities} Prior works on understanding dense loss landscapes have shown the existence of diverse low-loss solutions on the manifold of dense networks \citep{goodfellow2014qualitatively,garipov2018loss,draxler2018essentially,fort2019large}. Here, the fact that Selfish-RNN consistently achieves good performance with different runs naturally raises some questions: e.g., are final sparse connectivities obtained by different runs similar or very different? Is the distance between the original sparse connectivity and the final sparse connectivity large or small? To answer these questions, we investigate a method based on graph edit distance (GED) \citep{sanfeliu1983distance} to measure the topological distance between different sparse connectivities. The distance is scaled between 0 and 1. The smaller the distance is, the more similar the two sparse topologies are (See Appendix \ref{STDM} for details on how we measure the sparse connectivity distance).

The results are demonstrated in Figure \ref{fig:GED}. Figure \ref{fig:GED}-left shows how the topology of one random-initialized (random sparse connectivity and random weight values) network evolves when trained with Selfish-RNN. We compare the sparse connectivity at 5 epoch with those at the following epochs. We can see that the distance gradually increases from 0 to a very high value 0.8, meaning that Selfish-RNN optimizes the initial topology to a very different one. Moreover, Figure \ref{fig:GED}-right illustrates that the topological distance between two same-initialized (same sparse connectivity and same weight values) networks trained with different seeds after the $4^{th}$ epoch. Started from the same sparse topology, they evolve to completely different sparse topologies. While topologically different, they have very similarly performance. This indicates that in the case of RNNs there exist many low-dimensional sub-networks that can achieve similarly low loss. This phenomenon complements the findings of \citet{liu2020topological} which shows that there are numerous sparse sub-networks performing similarly well in the context of sparse MLPs.

% Moreover, ER usually causes RNN layers to be less sparse than other layers, resulting in a small increase of FLOPs.
% \textit{Erd{\H{o}}s-R{\'e}nyi}, introduced in SET, randomly sample the nonzero weights by the probability $P(M_{i,j}^l)=\min(\frac{\epsilon(n^l+n^{l-1})}{n^l\times{n^{l-1}}},1)$, where $\epsilon\in\mathbf{R^+}$ is a hyperparameter to control the sparsity level $S$ and $n^l$ is the number of neurons in the layer $l$. It distributes higher sparsity to the layers where $n^l$ is approximately in the same range with $n^{l-1}$, and  lower sparsity to the layers where $n^l\gg n^{l-1}$. Note that its variant, \textit{Erd{\H{o}}s-R{\'e}nyi-Kernel} , achieving better performance than uniform initialization with CNN models \citep{evci2019rigging}, scales as the original \textit{Erd{\H{o}}s-R{\'e}nyi} with RNN layers.

\subsection{Analysis of Weight Growth Methods} \label{ana_growth} Methods that leverage gradient-based weight growth (SNFS and RigL) have shown superiority on performance over the methods using random-based weight growth for CNNs \citep{evci2019rigging}. However, we observe a different behavior with RNNs. We set up a controlled experiment to compare these two methods with SNT-ASGD and momentum SGD. We report the results with various update intervals (the number of iterations between sparse connectivity updates) in Figure \ref{fig:com_gra_ran}. Surprisingly, gradient-based growth performs worse than random-based growth in most cases. Our hypothesis is that random growth helps in exploring better the search space, as it naturally considers a large number of various sparse connectivities during training, which is crucial to the performance of dynamic sparse training. Differently, gradient growth drives the network topology towards some similar local optima for the sparse connectivity as it uses a greedy search strategy (highest gradient magnitude) at every topological change. However, benefits provided by high-magnitude gradients might change dynamically afterwards due to complicated interactions between weights. We empirically illustrate our hypothesis via the proposed sparse connectivity distance measurement in Appendix \ref{TDGR}.

\subsection{Analysis of Sparse Initialization} It has been shown that the choice of sparse initialization (sparsity distribution) is important for sparse training in \citet{frankle2018the,kusupati2020soft,evci2019rigging}. Here, we compare two types of sparse initialization for RNNs, ER distribution and uniform distribution. Uniform distribution namely enforces the sparsity level of each layer to be the same as $S$. ER distribution allocates higher sparsity to larger layers than smaller ones. Note that its variant \textit{Erd{\H{o}}s-R{\'e}nyi-Kernel} proposed by \citet{evci2019rigging} scales back to ER for RNNs, as no kernels are involved. The results are shown as the \textit{Static} group in Table \ref{tab:stackedlstm}. We can see that uniform distribution outperforms ER distribution consistently for RNNs.

\subsection{Analysis of DST Hyperparameters} The sparsity $S$ and the initial pruning rate $p$ are two hyper-parameters of our DST method. We show their sensitivity analysis in Appendix \ref{ana_sparsity} and Appendix \ref{ana_pruning_rate}. We find that Selfish Stacked LSTMs, RHNs, ON-LSTM, and AWD-LSTM-MoS need around 25\%, 40\%, 45\%, and 40\% parameters to reach the performance of their dense counterparts, respectively. In addition, our method is quite robust to the choice of the initial pruning rate.
\section{Conclusion}
In this paper, we developed an approach to train sparse RNNs from scratch with a fixed parameter count throughout training. We further proposed SNT-ASGD, a specially designed sparse optimizer for training sparse RNNs and showed that it substantially improves the performance of all DST methods in RNNs. We observed that random-based growth achieves lower perplexity than gradient-based growth in the case of RNNs. Further, we developed an approach to compare two different sparse connectivities from the perspective of graph theory. Using this approach, we found that random-based growth explores better the topological search space for optimal sparse connectivities, whereas gradient-based growth is prone to drive the network towards similar sparse connectivity patterns, opening the path for a better understanding of sparse training.

% The sparse structure in our paper is enforced by binary masks, as off-the-shelf software and hardware have limited support for sparse operations. However, we want to highlight the fact that our method has promising potential to reduce memory and to boost the training process. We hope that our results will pile up on other researchers results in sparse training and soon there will be a change of perspective in such a way that the developers of deep learning software and hardware will start considering including real sparsity support in their solutions. 


\bibliography{example_paper}
\bibliographystyle{icml2021}

\clearpage

\appendix
\appendixpage

\section{Hyperparameters}
\label{app:Hyperparameters}
In this section, we share the hyperparameters used in thie paper. For fair comparison, we use the exact same hyperparameters and regularization introduced in ON-LSTM \citep{shen2018ordered} and AWD-LSTM-MoS \citep{yang2018breaking}.  We then extend the similar settings to stacked LSTMs and RHNs. No hyperparameter tuning techniques such as \citet{melis2017state} are involved in our experiments. No need of finetuning the original hyperparameters of the dense model is another advantage of our method. For all DST methods, the hyperparameters are the same, as shared in Table \ref{tab:hypo_hyper}.

\begin{table*}[ht!]
\tiny
\centering
\caption{Experiment hyperparameters including Optimizer (Opt), Learning rate (Lr), Batch size (Bs), Backpropagation through time (BPTT), Clip norm (Clip), Non-monotone interval for SNT-ASGD (Nonmono), Initial pruning rate (P); Lr Drop with (A, B) refers to B epochs with no improvement after which learning rate will be reduced by a factor of A; Dropout refers to the word-level dropout, embedding dropout, hidden layer dropout, and output dropout, respectively; Coupled means that the carry gate and the transform gate are coupled in RHNs; Tied means reusing the input word embedding matrix as the output matrix.}
\label{tab:hypo_hyper}
\begin{tabular}{lcccccccccccccc}
\toprule
Model & Data & Opt & Lr & Lr Drop & Bs & BPTT & Dropout & Epochs & Tied & Coupled & Clip & Nonmono & P\\
\midrule 
\multirow{3}{*}{Stacked LSTMs}& \multirow{3}{*}{PTB} &Adam & 0.001 & (2x, 2) &\multirow{3}{*}{20} &\multirow{3}{*}{35} &\multirow{3}{*}{(0, 0, 0.65, 0)} &\multirow{3}{*}{100} &\multirow{3}{*}{no} & \multirow{3}{*}{no}&\multirow{3}{*}{0.25} & \multirow{3}{*}{5} & \multirow{3}{*}{0.7}\\
 &  & SNT-ASGD & 40 & - &  &  &  &  &  &  &  & & \\
  &  & Momentum SGD & 2 & (1.33x, 1) &  &  &  &  &  &  &  & & \\
 \midrule 
\multirow{2}{*}{RHNs}& \multirow{2}{*}{PTB} &Adam & 0.001 & (2x, 2) &\multirow{2}{*}{20} &\multirow{2}{*}{35} &\multirow{2}{*}{(0.2, 0.65, 0.25, 0.65)} &\multirow{2}{*}{500} &\multirow{2}{*}{yes} & \multirow{2}{*}{yes}&\multirow{2}{*}{0.25} & \multirow{2}{*}{5} & \multirow{2}{*}{0.5}\\
 &  & SNT-ASGD & 15  & - &  &  &  &  &  &  &  & \\
 \midrule 
\multirow{2}{*}{ON-LSTM}& \multirow{2}{*}{PTB} &Adam & 0.001 & (2x, 2) &\multirow{2}{*}{20} &\multirow{2}{*}{70} &\multirow{2}{*}{(0.1, 0.5, 0.3, 0.45)} &\multirow{2}{*}{1000} &\multirow{2}{*}{yes} & \multirow{2}{*}{no}&\multirow{2}{*}{0.25} & \multirow{2}{*}{5} & \multirow{2}{*}{0.5}\\
 &  & SNT-ASGD & 30  & - &  &  &  &  &  &  &  & \\
  \midrule 
\multirow{2}{*}{AWD-LSTM-MoS}& \multirow{2}{*}{WikiText-2} &Adam & 0.001 & (2x, 2) &\multirow{2}{*}{15} &\multirow{2}{*}{70} &\multirow{2}{*}{(0.1, 0.55, 0.2, 0.4)} &\multirow{2}{*}{1000} &\multirow{2}{*}{yes} & \multirow{2}{*}{no}&\multirow{2}{*}{0.25} & \multirow{2}{*}{5} & \multirow{2}{*}{0.5}\\
 &  & SNT-ASGD & 15  & - &  &  &  &  &  &  &  & \\
\bottomrule
\end{tabular}
\end{table*}

\section{Ablation Study}
\label{app:ablation}
To analyze the influence of cell gate redistribution and Sparse NT-ASGD on the performance of sparse RNN training, we conduct an ablation study for all architectures. All models use the same hyperparameters with the ones reported in the main paper. We present the validation and testing perplexity for variants of our model without these two contributions, as shown in Table \ref{tab:ablation}. Not surprisingly, removing either of these two novelties degrades the performance. There is a significant degradation in the performance for all models, up to 13 perplexity point, if the optimizer switches to the standard NT-ASGD. This stands as empirical evidence regarding the benefit of SNT-ASGD. Without cell gate redistribution, the testing perplexity of all models degrades except for RHNs whose number of redistributed weights in each layer is only two. This indicates that cell gate redistribution is more effective for the models with more cell gates.

\begin{table*}[ht]
% \footnotesize
\centering
\caption{Ablation study of Selfish-RNN with stacked LSTMs, RHNs, ON-LSTM on Penn Treebank and AWD-LSTM-MoS on WikiText-2.}
\label{tab:ablation}
\begin{tabular}{l|c|c|c|c}
\toprule
Methods & Stacked LSTMs & RHNs & ON-LSTM & AWD-LSTM-MoS \\ 
\midrule
Selfish-RNN & 71.65 & 60.35 & 55.68 & 63.05 \\
\midrule
w/o cell gate redistribution & 72.89 & 60.26 & 57.48 & 65.27 \\
\midrule
w/o Sparse NT-ASGD & 73.74 & 69.70 & 69.28 & 71.65\\

\bottomrule
\end{tabular}
\end{table*}

\begin{table*}[ht]
\centering
\caption{A small experiment about the comparison among different cell gate redistribution methods. The experiment is evaluated with stacked LSTMs on Penn Treebank.}
\label{tab:ComGatesRedis}
\begin{tabular}{l|ccc}
\toprule
cell gate redistribution & \#Param & Validation & Test \\
\midrule
Mean of the magnitude of nonzero weights & 21.8M & 74.04 & 72.40 
\\
Mean of the gradient magnitude of nonzero weights & 21.8M  & 74.54 & 72.31    
\\
\midrule
Ours  & 21.8M & \textbf{73.76} & \textbf{71.65}     \\
\bottomrule
\end{tabular}
\end{table*}
\section{Comparison of Different Cell Gate Redistribution Methods}
\label{comparisonofAGR}

In Table \ref{tab:ComGatesRedis}, we conduct a small experiment to compare different methods of cell gate redistribution with stacked LSTMs. We consider weight redistribution based on the mean value of the magnitude of nonzero weights and the mean value of the gradient magnitude of nonzero weights. Our method can achieve the lowest perplexity.

% Considering the special architecture of LSTM, we align the input weight with the output from the previous layer and output of this layer, whereas, for the hidden weight, we align it with the hidden and output both from this layer. 

\section{Experimental Results with ON-LSTM}
\label{expre_onlstm}

\begin{table}[ht]
\footnotesize
\centering
\caption{Single model perplexity on validation and test sets for the Penn Treebank language modeling task with ON-LSTM. Methods indicated with ``ASGD'' are trained with SNT-ASGD. The numbers reported are averaged over five runs.}
\label{tab:ON-LSTM}
\renewrobustcmd{\bfseries}{\fontseries{b}\selectfont}
\renewcommand{\pm}{\mathbin{\mbox{\unboldmath$\mathchar"2206$}}}
\newcolumntype{L}[1]{>{\raggedright\let\newline\\\rraybackslash\hspace{0pt}}m{#1}}
\newcolumntype{C}[1]{>{\centering\let\newline\\\arraybackslash\hspace{0pt}}m{#1}}
\newcolumntype{R}[1]{>{\raggedleft\let\newline\\\arraybackslash\hspace{0pt}}m{#1}}
%\begin{tabular}{L{4cm}| C{0.8cm} S[table-figures-decimal = 1]
%S[table-figures-decimal = 1]}
\begin{tabular}{
  S[detect-weight, table-text-alignment=left, table-column-width=3.2cm, table-format=-2.2(1),mode=text]| 
  S[detect-weight, table-column-width=0.4cm, table-format=-2.2(1),mode=text]
  S[detect-weight, table-column-width=1.5cm, table-format=-2.2(1),mode=text]
  S[detect-weight, table-column-width=1.5cm, table-format=-2.2(1),mode=text]
}
\toprule
{Models} & \#Param & {Val} & {Test} \\ 
\midrule 
{Dense$_{1000}$ (NT-ASGD)} & {25M} & 58.29 \pm 0.10 & 56.17 \pm 0.12 \\
{Dense$_{1300}$ (NT-ASGD)} & {25M} & 58.55 \pm 0.11 &  56.28 \pm 0.19\\
{SET (Adam) } & {11.3M} & 65.90 \pm 0.08 & 63.56 \pm 0.14 \\
{DSR (Adam)} & {11.3M} & 65.22 \pm 0.07 &  62.55 \pm 0.06\\
{SNFS (Adam)} & {11.3M} & 68.00 \pm 0.10 & 65.52 \pm 0.15\\
{RigL (Adam)} & {11.3M} & 64.41 \pm 0.05 &  62.01 \pm 0.13\\
{RigL$_{1000}$ (ASGD) } & {11.3M} & 59.17 \pm 0.08 &  57.23 \pm 0.09\\
{RigL$_{1300}$ (ASGD) } & {11.3M} & 59.10 \pm 0.05 &  57.44 \pm 0.15\\
\midrule
{Selfish-RNN$_{1000}$ (ASGD)} & {11.3M} & 58.17 \pm 0.06 & 56.31 \pm 0.10 \\
{Selfish-RNN$_{1300}$ (ASGD)} & {11.3M} &  \bfseries 57.67 +- 0.03 & \bfseries 55.82 +- 0.11 \\
\bottomrule
    \end{tabular}
\end{table}

\begin{table}[ht]
\footnotesize
\centering
\caption{Single model perplexity on validation and test sets for the WikiText-2 language modeling task with AWD-LSTM-MoS. Baseline is AWD-LSTM-MoS obtained from ~\protect\citet{yang2018breaking}. Methods with ``ASGD'' are trained with SNT-ASGD.}
\label{tab:MOS}
% \begin{tabular}{p{1.8cm}|p{1cm}p{0.7cm}p{0.7cm}}
\begin{tabular}{l|ccc}
\toprule
Models & \#Param & Val & Test \\ 
\midrule
Dense (NT-ASGD)  & 35M & 66.01 & 63.33 \\
{SET (Adam)} & 15.6M & 72.82 & 69.61\\ 
{DSR (Adam)} & 15.6M & 69.95 &  66.93 \\
{SNFS (Adam)} & 15.6M & 79.97  & 76.18 \\
{RigL (Adam)} & 15.6M & 71.36 &  68.52 \\
{RigL (ASGD) } & {15.6M} & 68.84  &  65.18\\
\midrule
Selfish-RNN  (ASGD) & 15.6M & \textbf{65.96} & \textbf{63.05} \\
\bottomrule
\end{tabular}
\end{table}

% Further, the long-term information learned by high-ranking neurons will be shared with other time steps, whereas the information stored in low-ranking neurons will be replaced by the following input. 
Proposed by \citet{shen2018ordered} recently, ON-LSTM can learn the latent tree structure of natural language by learning the order of neurons. 
For a fair comparison, we use exactly the same model hyper-parameters and regularization used in ON-LSTM. We set the sparsity of each layer to 55\% and the initial pruning rate to 0.5. Same as ON-LSTM, we train the model for 1000 epochs and restart SNT-ASGD as a fine-tuning step once at the $500^{th}$ epoch, dubbed as Selfish-RNN$_{1000}$. As shown in Table \ref{tab:ON-LSTM}, Selfish-RNN outperforms the dense model while reducing the model size to 11.3M. Without SNT-ASGD, sparse training techniques can not reduce the test perplexity to 60. SNT-ASGD is able to improve the performance of RigL by 5 perplexity. Moreover, one interesting observation is that one of the regularizations used in the standard ON-LSTM, DropConnect, is perfectly compatible with our method, although it also drops the hidden-to-hidden weights out randomly during training.

In our experiments we observe that Selfish-RNN benefits significantly from the second fine-tuning operation. We scale the learning schedule to 1300 epochs with two fine-tuning operations at epoch 500 and 1000, respectively, dubbed as Selfish-RNN$_{1300}$. It is interesting that Selfish-RNN$_{1300}$ can achieve lower testing perplexity after the second fine-tuning step, whereas the dense model Dense$_{1300}$ can not even reach again the perplexity that it had before the second fine-tuning. The heuristic explanation here is that our method helps the optimization escape the local optima or a local saddle point by optimizing the sparse structure, while for dense models whose energy landscape is fixed, it is very difficult for the optimizer to find its way off the saddle point or the local optima.

\section{Experimental Results with AWD-LSTM-MoS}
\label{expre_mos}
We also evaluate Selfish-RNN on the WikiText-2 dataset. The model we choose is AWD-LSTM-MoS \citep{yang2018breaking}, which is the state-of-the-art RNN-based language model. It replaces Softmax with \textit{Mixture of Softmaxes} (MoS) to alleviate the Softmax bottleneck issue in modeling natural language. For a fair comparison, we exactly follow the model hyper-parameters and regularization used in AWD-LSTM-MoS. We sparsify all layers with 55\% sparsity except for the prior layer as its number of parameters is negligible. We train our model for 1000 epochs without finetuning or dynamical evaluation \citep{krause2018dynamic} to simply show the effectiveness of our method. As demonstrated in Table \ref{tab:MOS}, Selfish AWD-LSTM-MoS can match the performance of the corresponding dense model with 15.6M parameters.

\begin{figure*}[h]
\centering
\begin{subfigure}[b]{0.45\textwidth}
  \includegraphics[width=7cm, height=5cm]{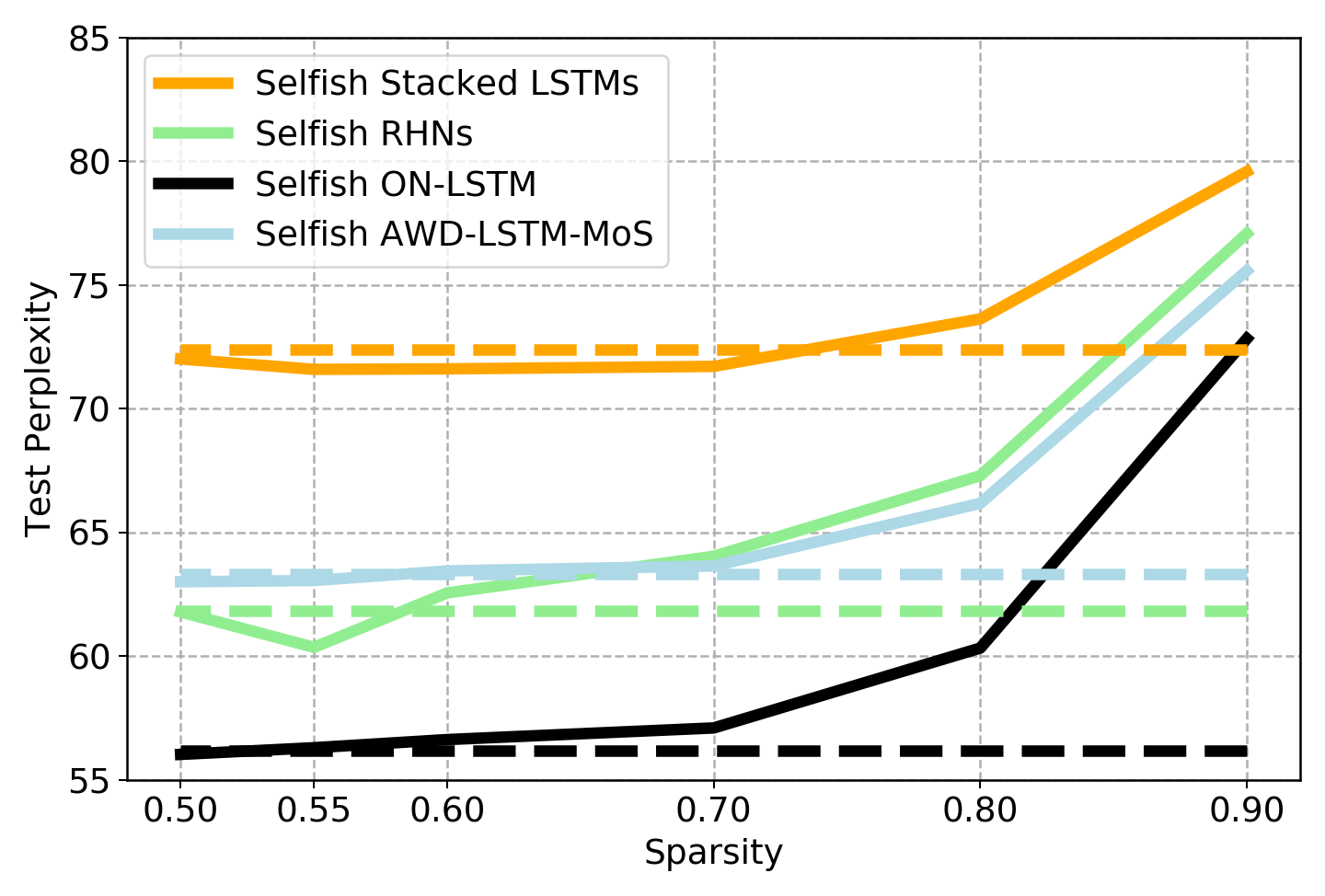}
  \caption{Performance Sensitivity of Sparsity}\label{fig:sparsity}
\end{subfigure}
\hspace{0.2cm}
\begin{subfigure}[b]{0.45\textwidth}
  \includegraphics[width=7cm, height=5cm]{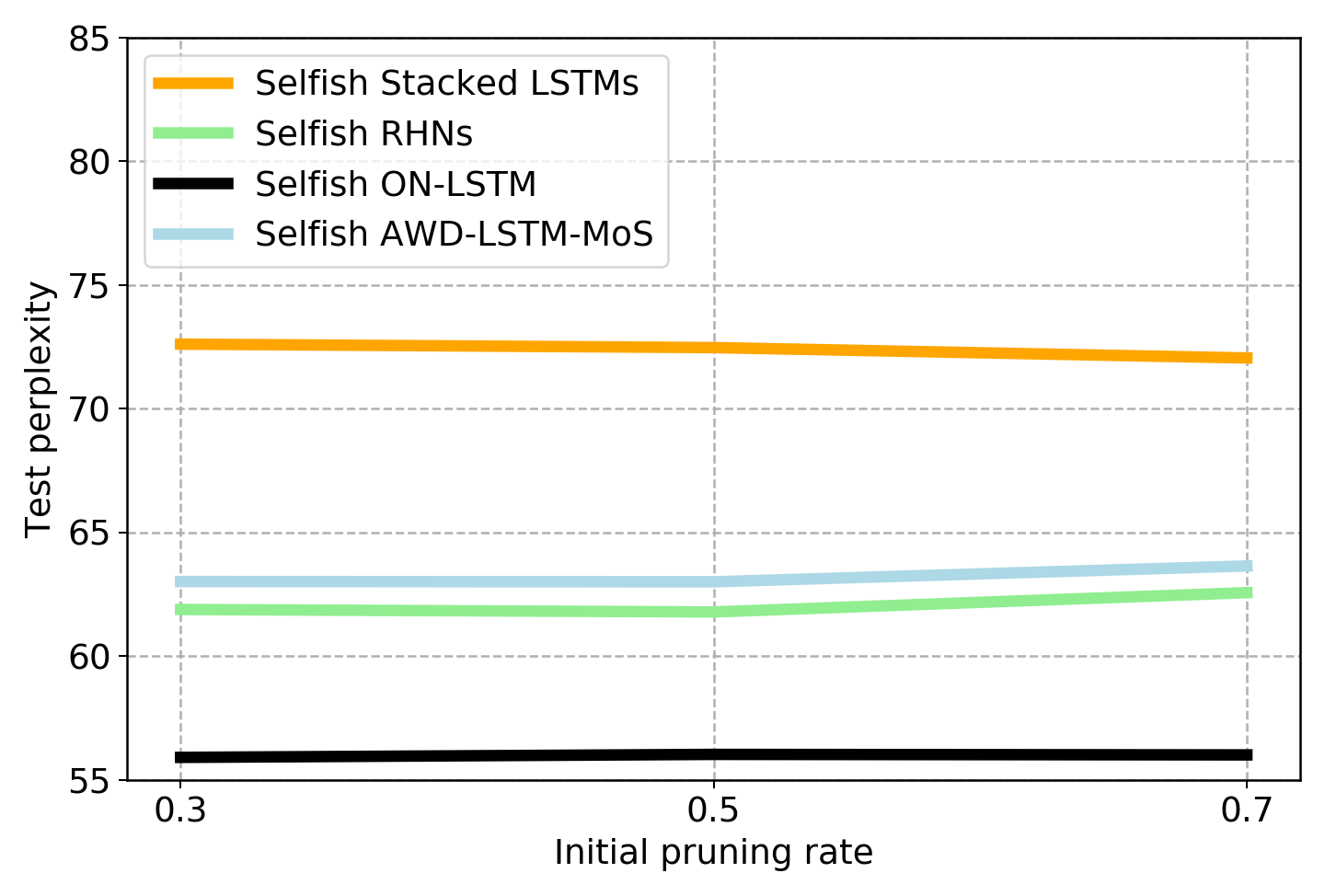}
  \caption{Performance Sensitivity of Pruning Rate}\label{fig:pruning_rate}
\end{subfigure}

\caption{Sensitivity analysis of sparsity levels $S$ and initial pruning rates $p$ with Selfish stacked LSTMs, RHNs, ON-LSTM, and AWD-LSTM-MoS. (a) Test perplexity of all models with various sparsity levels. The initial pruning rate is 0.7 for stacked LSTMs, and 0.5 for the rest models. The dashed lines represent the performance of the corresponding dense models. (b) Test perplexity of all models with different initial pruning rates. The sparsity level is 67\%, 52.8\%, 55\% and 55\% for Selfish stacked LSTMs, RHNs, ON-LSTM, and AWD-LSTM-MoS, respectively.}
\label{fig:sparsity_pruning}
\end{figure*}

\section{Effect of Sparsity}
\label{ana_sparsity}
There is a trade-off between the sparsity level $S$ and the test perplexity of Selfish-RNN. When there are too few parameters, the sparse neural network will not have enough capacity to fit the data. Here, we analyze this trade-off by training all models with Selfish-RNN at various sparsity levels $S \in [0.50, 0.55, 0.60, 0.70, 0.80, 0.90]$, reported in Figure \ref{fig:sparsity}. We find that Selfish Stacked LSTMs, RHNs, ON-LSTM, and AWD-LSTM-MoS need around 25\%, 40\%, 45\%, and 40\% parameters to reach the performance of their dense counterparts, respectively. 

\section{Effect of Initial pruning rate}
\label{ana_pruning_rate}
The initial pruning rate $p$ determines how many weights would be removed at each connectivity update. We analyze the performance sensitivity of our algorithm to the initial pruning rate $p$ by varying it $\in \left[ 0.3, 0.5, 0.7 \right]$. We set the sparsity level of each model as the one having the best performance in Figure \ref{fig:sparsity}. Results are shown in Figure \ref{fig:pruning_rate}. We can clearly see that our method is very robust to the choice of the initial pruning rate.

\section{Difference Among SET, Selfish-RNN and Iterative Pruning Methods}
\label{diff_set_RNN_pruning}
The topology update strategy of Selfish-RNN differs from SET in several important features (1) we automatically redistribute weights across cell gates for better regularization, (2) we use magnitude-based removal instead of removing a fraction of the smallest positive weights and the largest negative weights, (3) we use uniform initialization rather than non-uniform sparse distribution like ER or ERK. Additionally, the optimizer proposed in this work, SNT-ASGD, brings substantial perplexity improvement to the sparse RNN training.

Iterative pruning and retraining techniques \citep{han2015deep,zhu2017prune,frankle2018the} usually involve three steps (1) pre-training a dense model, (2) pruning the unimportant based on some criteria, and (3) re-training the pruned model to improve performance. The pruning and re-training cycle is required at least once, but may many times depending on the specific algorithms used. Therefore, the computational costs required by iterative pruning and retraining is at least the same as fully training a dense model. Different from iterative pruning and retraining, FLOPs required by Selfish-RNN is proportional to the density of the model, as it allows us to train a sparse network with a fixed number of parameters throughout training in one single run, without any retraining phases. Moreover, the overhead caused by the dynamic sparse connectivity operation is negligible, as it performs only once per epoch.

\section{Comparison between Selfish-RNN and Pruning}
\label{com_pruning}
\begin{figure}[ht!]
\begin{center}
\includegraphics[width=7cm, height=5cm]{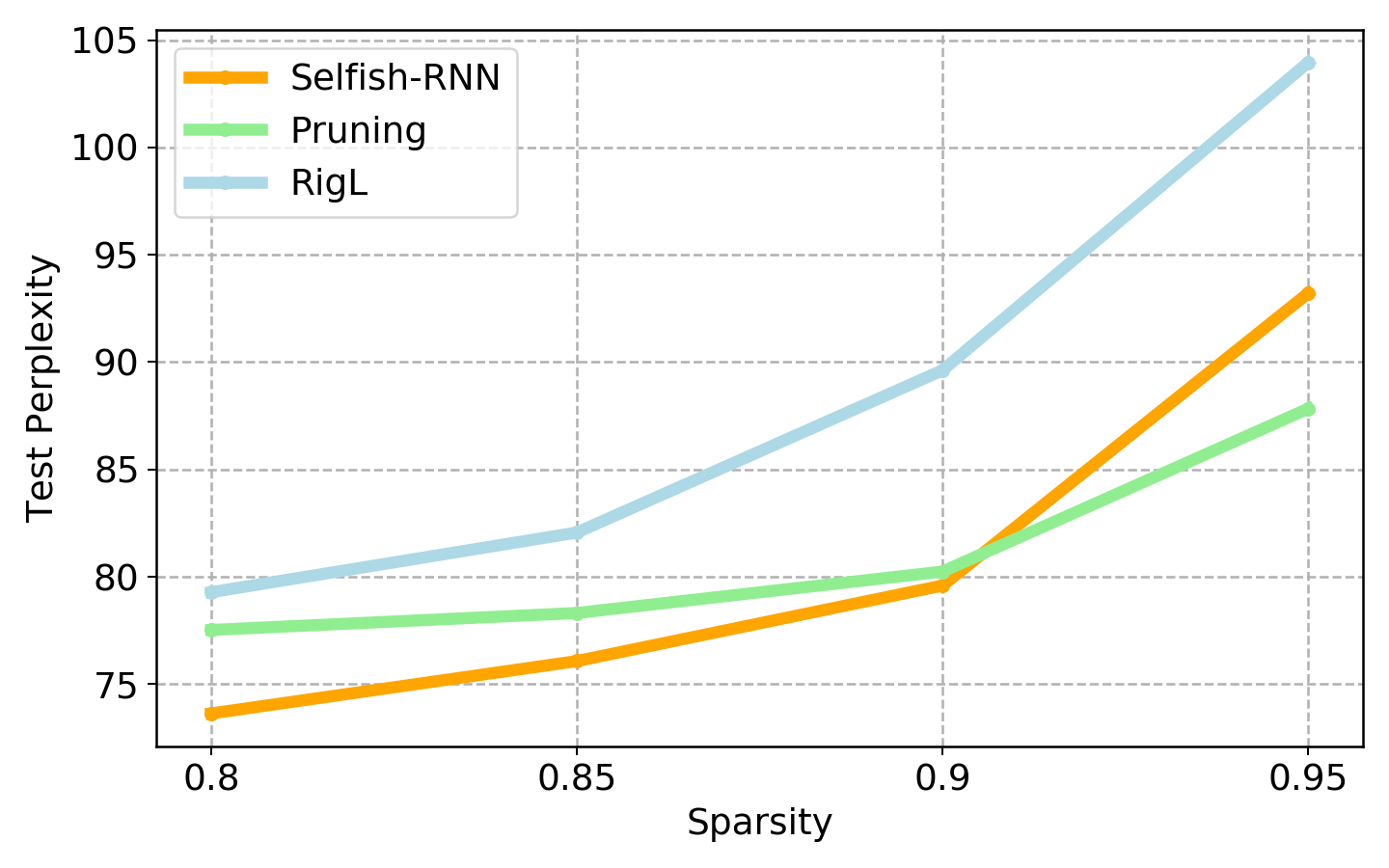}
\caption{Comparison among Selfish-RNN, RigL and iterative magnitude pruning with stacked LSTMs on PTB. The pruning baseline is obtained from \citet{zhu2017prune}.}
\label{Fig:Effi_pruning}
\end{center}
\end{figure}

It has been shown by \citet{evci2019rigging} that while state-of-the-art sparse training method (RigL) achieves promising performance with various CNN models, it fails to match the performance of pruning in RNNs. Given the fact that magnitude pruning has become a widely-used and strong baseline for model compression, we also report a comparison between Selfish-RNN and iterative magnitude pruning with stacked LSTMs. The pruning baseline is obtained from \citet{zhu2017prune}. The results are demonstrated in Figure \ref{Fig:Effi_pruning}-right.

We can see that Selfish-RNN exceeds the performance of pruning in most cases. An interesting phenomenon is that, with increased sparsity, we see a decreased performance gap between Selfish-RNN and pruning. Especially, Selfish-RNN performs worse than pruning when the sparsity level is 95\%. This can be attributed to the poor trainability problem of sparse models with extreme sparsity levels. Noted in \citet{Lee2020A}, the extreme sparse structure can break dynamical isometry \citep{saxe2013exact} of sparse networks, which subsequently degrades the trainability of sparse neural networks. Different from sparse training methods, pruning operates from a dense network and thus, does not have this problem.

\begin{figure*}[ht!]
\hspace{0.8cm}
\begin{subfigure}[b]{0.45\textwidth}
  \includegraphics[width=8cm, height=5cm]{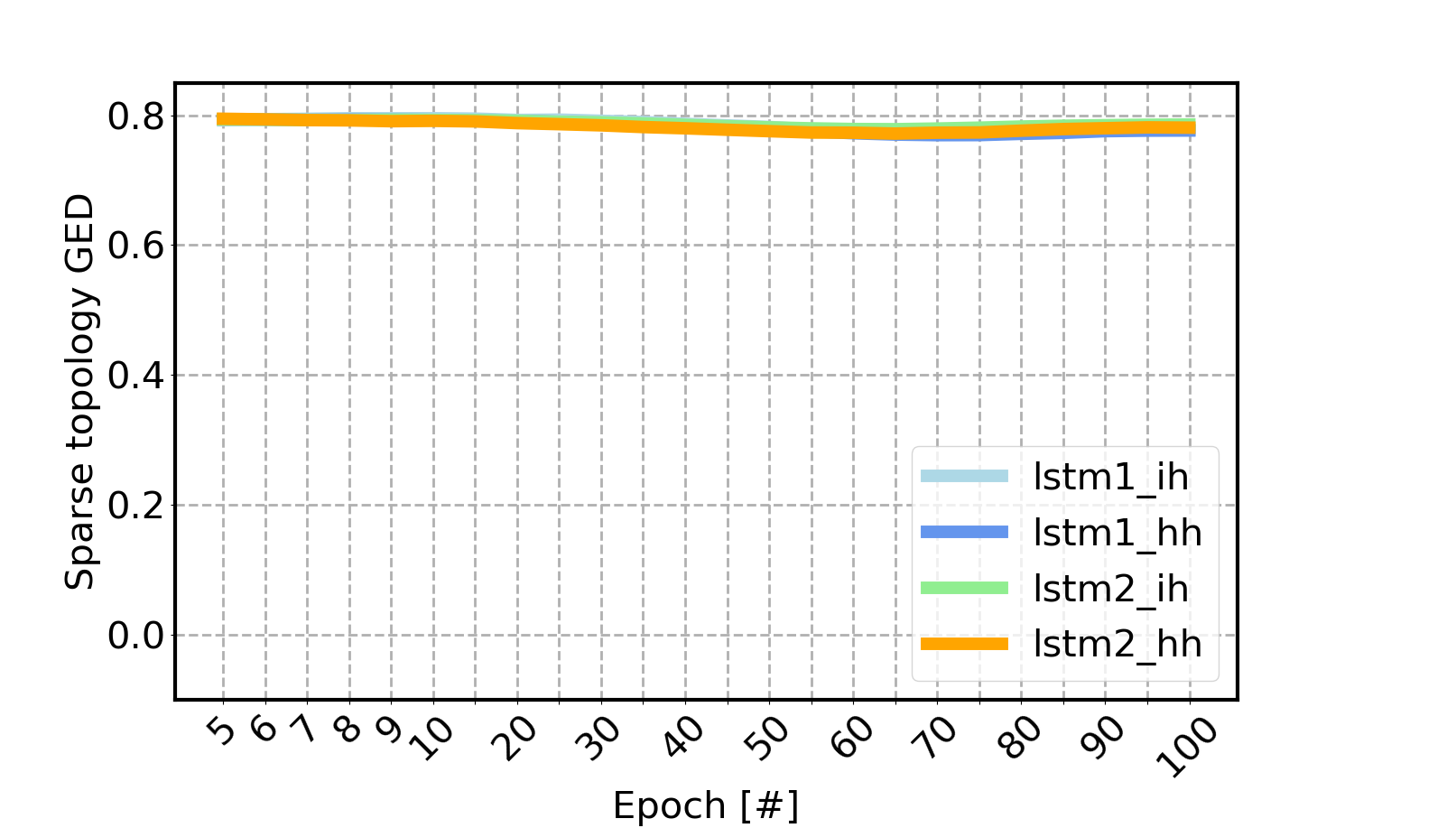}
%   \caption{Sparsity sensitivity}\label{fig:sparsity}
\end{subfigure}
\hspace{0.2cm}
\begin{subfigure}[b]{0.45\textwidth}
  \includegraphics[width=8cm, height=5cm]{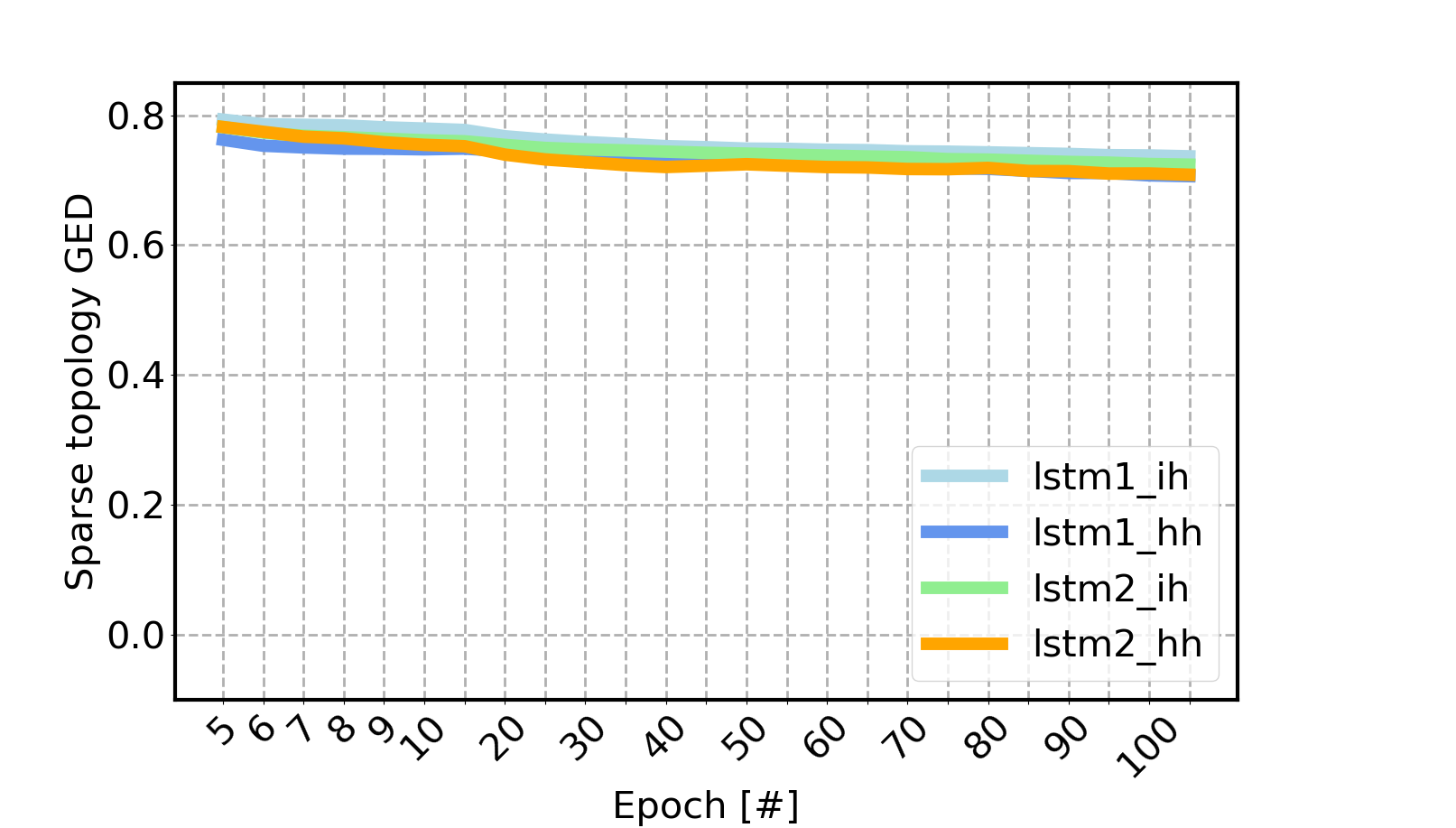}
%   \caption{pruning rate sensitivity}label{fig:pruning_rate}
\end{subfigure}
\caption{(\textbf{left}) The sparse connectivity distance between two different training runs of stacked LSTMs trained with random growth. (\textbf{right}) The sparse connectivity distance between two different training runs of stacked LSTMs trained with gradient growth. }
\label{Fig:TDGR_switch}
\end{figure*}

\begin{figure*}[ht!]
\hspace{0.8cm}
\begin{subfigure}[b]{0.45\textwidth}
  \includegraphics[width=8cm, height=5cm]{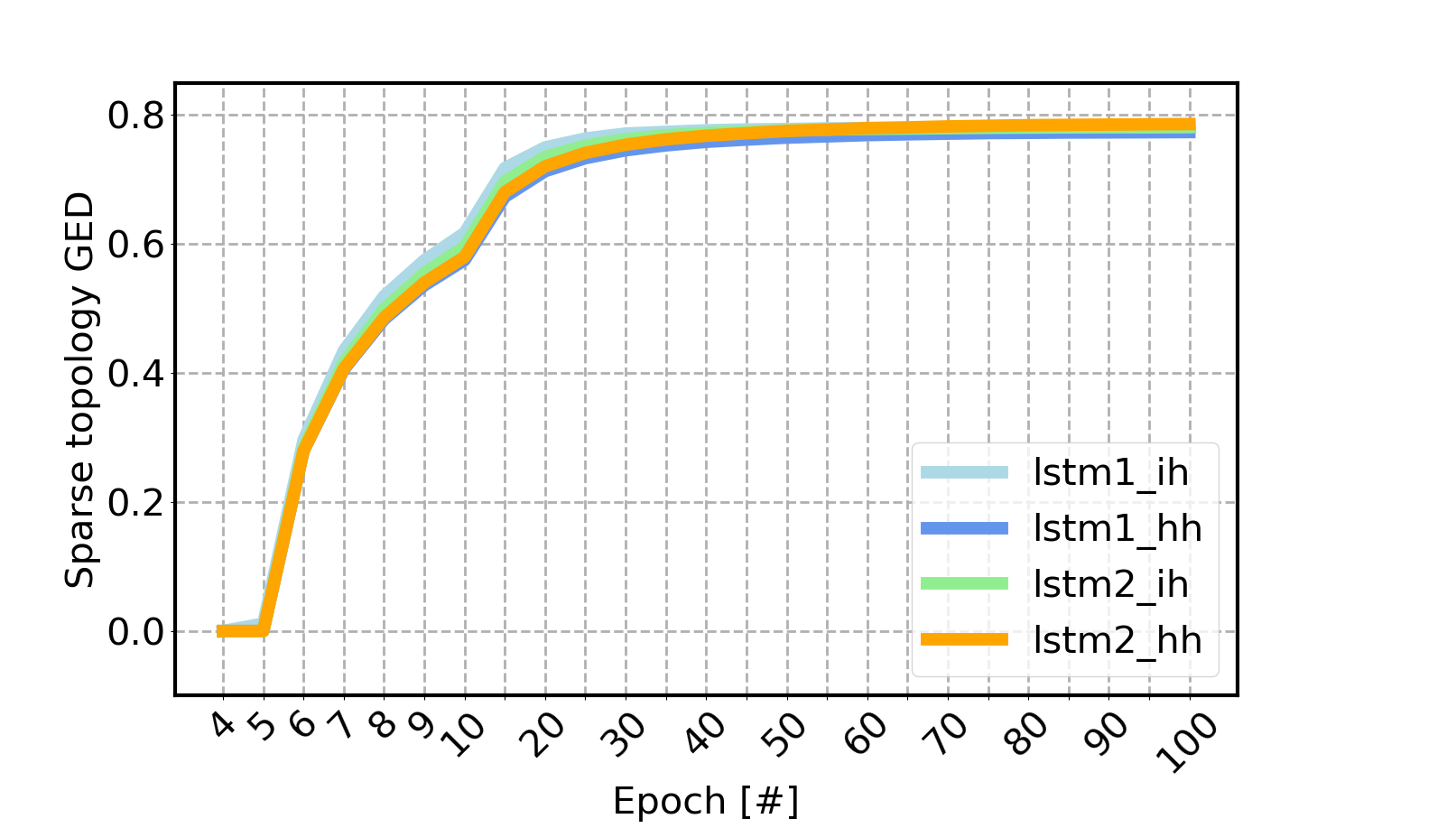}
%   \caption{Sparsity sensitivity}\label{fig:sparsity}
\end{subfigure}
\hspace{0.2cm}
\begin{subfigure}[b]{0.45\textwidth}
  \includegraphics[width=8cm, height=5cm]{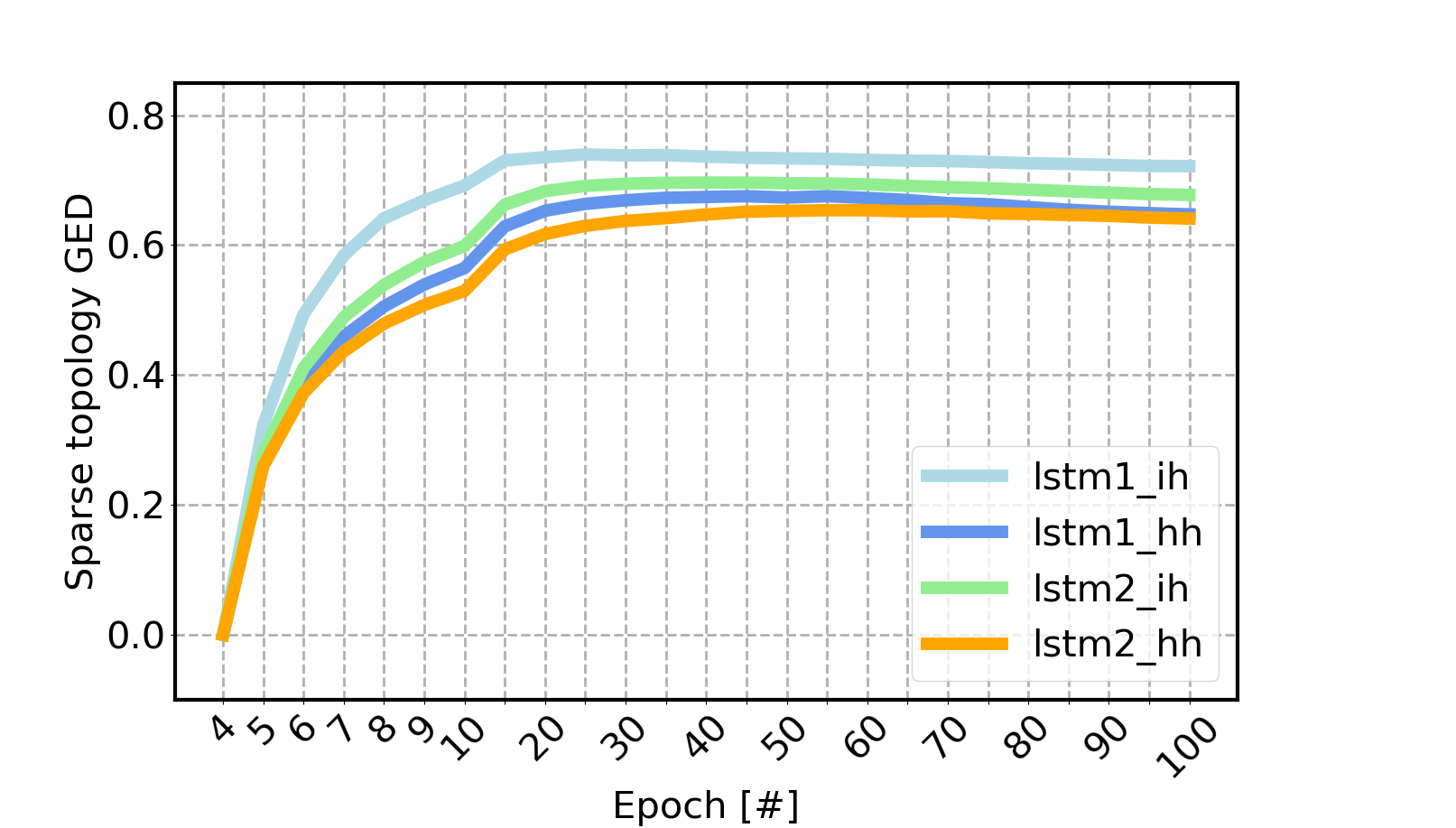}
%   \caption{pruning rate sensitivity}label{fig:pruning_rate}
\end{subfigure}
\caption{(\textbf{left}) The sparse connectivity distance between two stacked LSTMs with same initialization but different training seeds trained with random growth. (\textbf{right}) The sparse connectivity distance between two networks with same initialization but different training seeds trained with gradient growth. $ih$ is the input weight tensor comprising four cell gates and $hh$ is the hidden state weight tensor comprising four cell gates.}
\label{Fig:TDGR_different}
\end{figure*}
\section{Sparse Connectivity Distance Measurement}
\label{STDM}
Our sparse connectivity distance measurement considers the unit alignment based on a \textit{semi-matching} technique introduced by \citet{li2016convergent} and a graph distance measurement based on graph edit distance (GED) \citep{sanfeliu1983distance}. More specifically, our measurement includes the following steps: 
\begin{enumerate}
  \item  We train two sparse networks with dynamic sparse training on the training dataset and store the sparse topology after each epoch. Let ${\rm \boldsymbol{W}}_l^i$ be the set of sparse topologies for the $l$-th layer of network $i$. 
  \item Using the saved model, we compute the activity output on the test data, ${\rm \boldsymbol{O}}_l^i \in \mathbb{R}^{n\times{m}}$, where $n$ is the number of hidden units and $m$ is the number of samples.
  \item We pair-wisely match two topologies obtained from different networks ${\rm \boldsymbol{W}}_l^i$ and ${\rm \boldsymbol{W}}_l^j$ by the semi-matching method introduced in \citet{li2016convergent} based on their activity units. The semi-matching step is achieved by finding the a pair of units from different networks with the maximum correlation. 
  \item After alignment, we apply graph edit distance (GED) to measure the similarity between ${\rm \boldsymbol{W}}_l^i$ and ${\rm \boldsymbol{W}}_l^j$. Eventually, the distance is scaled to lie between 0 and 1. The smaller the distance is, the more similar the two sparse topologies are.
\end{enumerate}

Here, we choose stacked LSTMs on PTB dataset as a specific case to analyze. Specifically, we train two stacked LSTMs for 100 epochs with different random seeds. We choose a relatively small pruning rate of 0.1. We start alignment at the $5^{th}$ epoch to ensure a good alignment, as the model does not learn very well at the very beginning of training.

\section{Sparse Connectivity Distance Comparison between Different Growth Methods}
\label{TDGR}

In this section, we investigate the topological distance between sparse connectivities learned by gradient weight growth and random weight growth. We empirically illustrate that gradient growth drives different networks into some similar connectivity patterns based on the proposed distance measurement between sparse connectivities. The initial pruning rates are set as 0.1 for all training runs in this section. 

First, we measure the sparse connectivity distance between two different training runs trained with gradient growth and random growth, respectively, as shown in Figure \ref{Fig:TDGR_switch}. We can see that, starting with very different sparse connectivity topologies, two networks trained with random growth end up at the same distance, whereas the distance between two networks trained with gradient growth is continuously decreasing and this tendency is likely to continue as the training goes on. We further report the distance between two networks with same initialization (same sparse connectivity and same weight values) but different training seeds in Figure \ref{Fig:TDGR_different}. We can see that the distance between sparse connectivities optimized by gradient growth is smaller than the ones optimized by random growth.

These observations are in line with our hypothesis and indicate that gradient growth drives networks into some similar structures, whereas random growth allows models to explore more sparse structures spanned over the dense network, and thus has a better chance to find a better sparse connectivity.

\section{FLOPs Analysis of Different Approaches}
\label{app:training_flops}

\begin{table*}[ht]
\vspace{-0.1cm}
% \footnotesize
\caption{Single model perplexity on validation and test sets for the Penn Treebank language modeling task with stacked LSTMs and RHNs. FLOPs required to train the entire model and to test on single sample are reported. ‘*’ indicates the reported results from the original papers: ``Dense'' is obtained from \citet{zaremba2014recurrent} and ISS is obtained from \citet{wen2018learning}. ``Static-ER'' and ``Static-uni'' are the static sparse network trained from scratch with ER distribution and uniform distribution, respectively. ``Small'' refers the small-dense network.}
\label{tab:stackedlstm_flops}
% \begin{tabular}{p{3cm}| p{1.2cm}|p{1.2cm}|p{0.58cm}| p{0.58cm} || p{1.2cm}|p{1.2cm}|p{0.58cm} | p{0.58cm}}
\begin{tabular}{l|cccc||cccc}
\toprule
& \multicolumn{4}{c||}{Stacked LSTMs} &  \multicolumn{4}{c}{RHNs}\\
\midrule
Models & FLOPs & FLOPs & Val & Test & FLOPs & FLOPs & Val & Test \\
&  (Train) & (Test)&  &   & (Train) & (Test)&  &  \\
\midrule
Dense* & 1x(3.1e16)  & 1x(7.2e10) &82.57 & 78.57 & 1x(6.5e16) & 1x(3.3e10) & 67.90 & 65.40\\
% & (3.1e16) & (7.2e10) & & &  (6.5e16) & (3.3e10) & &\\
Dense (NT-ASGD)  & 1x & 1x & 74.51 & 72.40  & 1x & 1x & 63.44 & 61.84\\
\midrule
\midrule
& \multicolumn{4}{c||}{S=0.67} &  \multicolumn{4}{c}{S=0.53}\\
\midrule
Small (NT-ASGD) & 0.33x & 0.33x & 88.67 & 86.33  & 0.47x & 0.47x & 70.10 & 68.40  \\
Static-ER (SNT-ASGD)& 0.33x & 0.34x & 81.02 & 79.30  & 0.47x & 0.47x & 75.74 & 73.21 \\
Static-uni (SNT-ASGD)& 0.33x & 0.33x & 80.37 & 78.61 & 0.47x & 0.47x & 74.11 & 71.83 \\
\midrule

ISS*  & 0.28x & 0.20x & 82.59 & 78.65 & 0.50x & 0.47x & 68.10 & 65.40 \\
GMP (Adam)  & 0.63x & 0.33x & 89.47 & 87.97 & 0.62x & 0.47x & 63.21 & 61.55 \\
SET (Adam)  & 0.33x & 0.34x & 87.30 & 85.49 & 0.47x & 0.47x & 63.66 & 61.08 \\
DSR (Adam)  & 0.38x & 0.40x & 89.95 & 88.16 & 0.47x & 0.47x & 65.38 & 63.19 \\
SNFS (Adam)  & 0.63x & 0.38x & 88.31 & 86.28  & 0.63x & 0.45x & 74.02 & 70.99 \\
RigL (Adam)  & 0.33x & 0.34x & 88.39 & 85.61  & 0.47x & 0.47x & 67.43 & 64.41 \\
Selfish-RNN (Adam) & 0.33x & 0.33x & 85.70 & 82.85 & 0.47x & 0.47x & 63.28 & 60.75\\
GMP (SNT-ASGD)  & 0.63x & 0.33x & 76.78 & 74.84 & 0.62x & 0.47x & 65.63 & 63.96 \\
RigL (SNT-ASGD) & 0.33x & 0.34x & 78.31 & 75.90 & 0.47x & 0.47x &  64.82 & 62.47 \\
Selfish-RNN (SNT-ASGD) & 0.33x & 0.33x &  \textbf{73.76} &  \textbf{71.65}  & 0.47x & 0.47x & \textbf{62.10} & \textbf{60.35}  \\
\midrule
\midrule
& \multicolumn{4}{c||}{S=0.62} &  \multicolumn{4}{c}{S=0.68}\\
\midrule
ISS*  & 0.32x & 0.23x & 80.24 & 76.03  & 0.34x & 0.32x & 70.30 & 67.70  \\
% GMP  & 0.66x & 0.38x & 89.28 & 87.79  & 0.51x & 0.32x & 63.22 &  61.39 \\
GMP (SNT-ASGD)  & 0.63x & 0.38x & 74.86 & 73.03 & 0.51x & 0.32x & 66.61 & 64.98 \\
RigL (SNT-ASGD)   & 0.38x & 0.39x & 77.16 & 74.76  & 0.32x & 0.32x & 69.32 & 66.64 \\
Selfish-RNN (SNT-ASGD) & 0.38x & 0.38x &\textbf{73.50} & \textbf{71.42}  & 0.32x & 0.32x & \textbf{66.35} & \textbf{64.03}  \\
% DSR (SNT-ASGD)  & 25.2M & 0.380x & 0.376x & 72.98 & 71.36 \\
\bottomrule
\end{tabular}
\end{table*}

\begin{table*}[ht!]
\centering
\caption{Training FLOPs analysis of different sparse training approaches. $f_D$ refers to the training FLOPs for a dense model to compute one single prediction in the \textit{forward pass} and $f_S$ refers to the training FLOPs for a sparse model. $\mathit{\Delta}T$ is the number of iterations used by RigL to update sparse connectivity. $s_t$ is the sparsity level of the model at iteration $t$.}
\label{tab:FLOPS}
\begin{tabular}{l|cccc}
\toprule
Method & Forward Pass & Backward Pass &  Total  \\ 
\midrule
Dense &  $f_D$ & $2f_D$ & $3f_D$ \\
ISS &  $f_D*s_t$ & $2f_D*s_t$ & $3f_D*s_t$ \\
Pruning & $f_D*s_t$ & $2f_D*s_t$ & $3f_D*s_t$ \\
SET &  $f_S$ & $2f_S$ & $3f_S$  \\
DSR & $f_S$ & $2f_S$ & $3f_S$ \\
SNFS & $f_S$ & $f_S$ + $f_D$ & $2f_S$ +$f_D$ \\
RigL & $f_S$ & $\frac{(2\mathit{\Delta}T+1)f_S+f_D}{\mathit{\Delta}T+1}$ & $\frac{3f_S\mathit{\Delta}T+2f_S+f_D}{\mathit{\Delta}T+1}$ \\
Selfish-RNN (ours) &  $f_S$ & $2f_S$ & $3f_S$  \\
\bottomrule
\end{tabular}
\end{table*}

Although different DST methods maintain a fix parameter count throughout training, their training costs can be very different since different sparse distributions lead to different computational costs. Hence, we also report the estimated training and inference FLOPs for all methods in this section. 

% As off-the-shelf software and hardware have limited support for sparse operations, we 

We follow the way of calculating training FLOPs proposed by \citet{evci2019rigging}. The perplexity and the corresponding training and inference FLOPs of different methods are given in Table \ref{tab:stackedlstm_flops}.  We split the process of training a sparse recurrent neural network into two steps: \textit{forward pass} and \textit{backward pass}.

\paragraph{\textit{Forward pass}} In order to calculate the loss of the current models given a batch of input data, the output of each layer is needed to be calculated based on a linear transformation and a non-linear activation function. Within each RNN layer, different cell gates are used to regulate information in sequence using the output of the previous time step and the input of this time step.  

\paragraph{\textit{Backward pass}} In order to update weights, during the backward pass, each layer calculates 2 quantities: the gradient of the loss function with respect to the activations of the previous layer and the gradient of the loss function with respect to its own weights. Therefore, the computational expense of \textit{backward pass} is twice that of \textit{forward pass}. Given that RNN models usually contain an embedding layer from which it is very efficient to pick a word vector, for models not using weight tying, we only count the computations to calculate the gradient of its parameters as the training FLOPs and we omit its inference FLOPs. For models using weight tying, both the training FLOPs and the inference FLOPs are omitted. 

Given a specific architecture, we denote $f_D$ as dense FLOPs required to finish one training iteration and $f_S$ as the corresponding sparse FLOPs ($f_S \approx (1-S)f_D$), where $S$ is the sparsity level. Thus $ f_S \ll f_D $ for very sparse networks.  Since different sparse training methods use different sparse distribution, their FLOPs $f_S$ are also different from each other. We omit the FLOPs used to update the sparse connectivity, as it is only performed once per epoch. Overall, the total FLOPs required for one training update on one single sample are given in Table \ref{tab:FLOPS}. The training FLOPs of dense-to-sparse methods like, ISS and pruning, are $3f_D*s_t$, where $s_t$ is the sparsity of the model at iteration $t$. Since dense-to-sparse methods require to train a dense model for a while, their training FLOPs and memory requirement are higher than our method. For methods that allow the sparsity of each layer dynamically changing e.g., DSR and SNFS, we approximate their training FLOPs via their final distribution, as their sparse distribution converges to the final distribution in the first few epochs. ER distribution causes a bit more inference FLOPs than uniform distribution because it allocates more weights to the RNN layers than other layers. SNFS requires extra FLOPs to calculate dense momentum during the backward pass. Although RigL also uses the dense gradients to assist weight growth, it only needs to calculate dense gradients every $\mathit{\Delta}T$ iterations, thus with a smaller number of FLOPs given by $\frac{3f_S\mathit{\Delta}T+2f_S+f_D}{\mathit{\Delta}T+1}$. Here, we simply omit the extra FLOPs required by the full gradient calculation as it is negligible compared with the whole training FLOPs. Moreover, the inference FLOPs are calculated with the final sparse distribution learned by different methods.

\section{Final Cell Gate Sparsity Breakdown}
\label{breakdown}
\begin{figure*}[ht]
\begin{center}
\centerline{\includegraphics[width=0.7\textwidth]{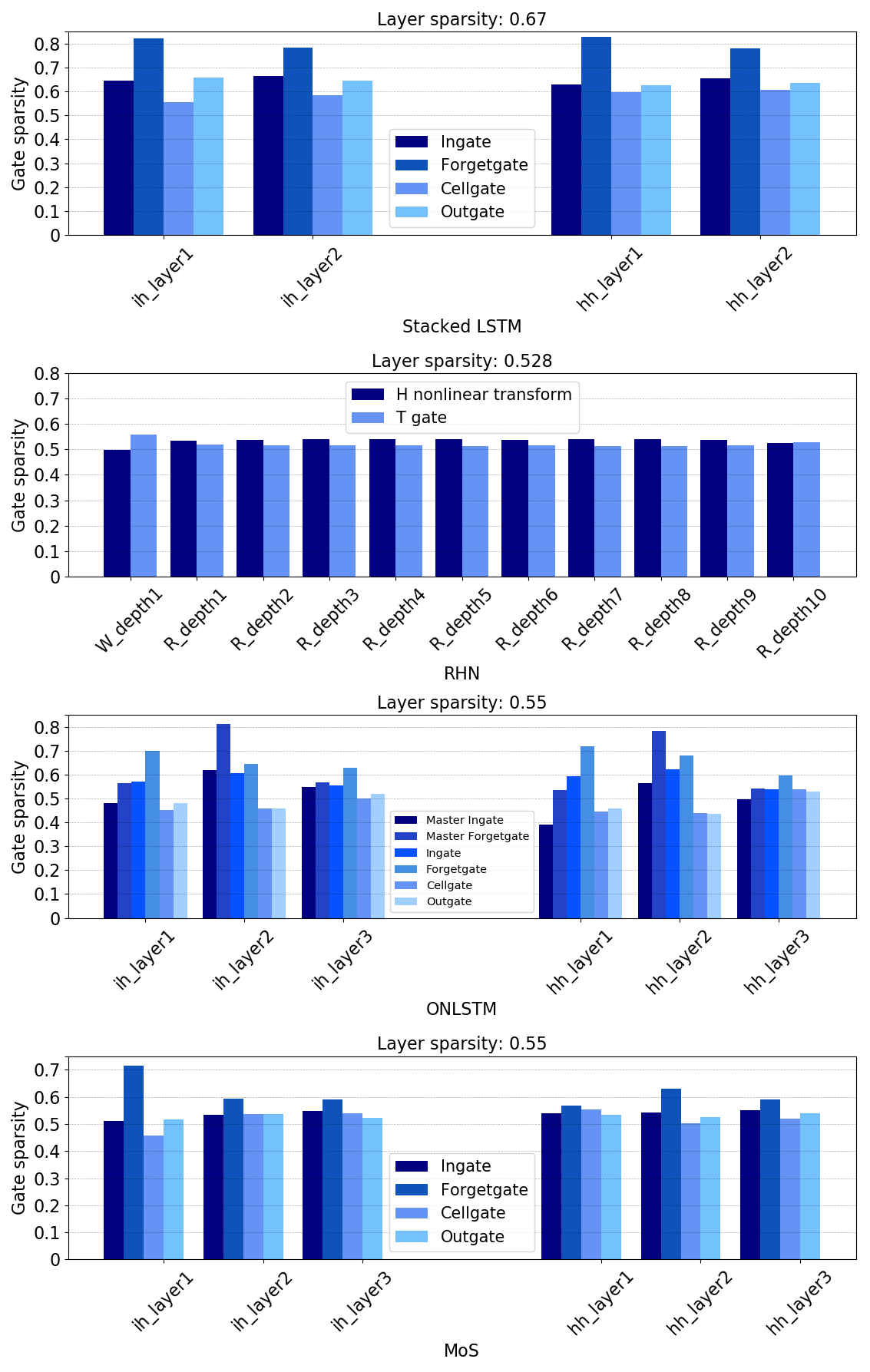}}
\caption{Breakdown of the final sparsity level of cell gates with stacked LSTMs, RHNs, ON-LSTM on PTB, AWD-LSTM-MoS on Wikitext-2. $W$ and $R$ is the weight of the H nonlinear transform and the T gate in RHNs, respectively; $ih$ and $hh$ refer to the input weight and the hidden weight of each LSTM layer, respectively.} 

\label{fig:breakdown}
\end{center}
\end{figure*}

We further investigate the final sparsity level of different cell gates learned automatically by our method in Figure \ref{fig:breakdown}. We find a consistent observation existing in all models, i.e., the weight of the forget gates, either the forget gate in the standard LSTM or the master forget gate in ON-LSTM, tend to be sparser than other gate weights. The weight of the cell gates and output gates are denser than the average. However, there is no big difference between the gates in RHNs, even although the $H$ nonlinear transform gate is slightly sparser than the $T$ gate weight in most RHNs layers. 

\section{Limitation}
The aforementioned training benefits have not been fully explored, as off-the-shelf software and hardware have limited support for sparse operations. The unstructured sparsity is difficult to be efficiently mapped to the existing parallel processors. The results of our paper provide motivation for new types of hardware accelerators and libraries with better support for sparse neural networks. Nevertheless, many recent works have been developed to accelerate sparse neural networks including \citet{gray2017gpu,moradi2019sparsemaps,ma2019transformed,yang2019feed,liu2020sparse}. For instance, NVIDIA develops the A100 GPU enabling the Fine-Grained Structured Sparsity \citep{NVIDIA2020}. The sparse structure is enforced by allowing two nonzero values in every four-entry vector to reduce memory storage and bandwidth by almost $2\times$. We hope that our results will pile up on other researchers results in sparse training and soon there will be a change of perspective in such a way that the developers of deep learning software and hardware will start considering including real sparsity support in their solutions. 
%%%%%%%%%%%%%%%%%%%%%%%%%%%%%%%%%%%%%%%%%%%%%%%%%%%%%%%%%%%%%%%%%%%%%%%%%%%%%%%
%%%%%%%%%%%%%%%%%%%%%%%%%%%%%%%%%%%%%%%%%%%%%%%%%%%%%%%%%%%%%%%%%%%%%%%%%%%%%%%

\end{document}